\documentclass[pmlr,twocolumn,10pt]{jmlr} 

\usepackage{booktabs}
\usepackage{siunitx}
\usepackage{soul}
\usepackage{multirow}
\usepackage{arydshln}
\usepackage{caption}
\usepackage{graphicx}
\usepackage{subcaption}
\usepackage{xcolor}
\definecolor{customgray}{gray}{0.9}
\usepackage[many]{tcolorbox}    	
\usepackage{setspace}               
\usepackage{multicol}               
\usepackage{algorithm2e}
\usepackage{xcolor, soul}

\usepackage[hang,flushmargin]{footmisc}
\usepackage{lipsum}
\makeatletter
\newcommand{\algorithmfootnote}[2][\footnotesize]{%
  \let\old@algocf@finish\@algocf@finish
  \def\@algocf@finish{\old@algocf@finish
    \leavevmode\rlap{\begin{minipage}{\linewidth}
    #1#2
    \end{minipage}}%
  }%
}

\definecolor{main}{HTML}{000000}    

\newtcolorbox{boxB}{
    fontupper = \color{main}, 
    boxrule = 1.5pt,
    colframe = main,
    rounded corners,
    arc = 5pt   
}
\usepackage[switch]{lineno}

\theorembodyfont{\upshape}
\theoremheaderfont{\scshape}
\theorempostheader{:}
\theoremsep{\newline}

\makeatletter
\renewcommand{\@fnsymbol}[1]{%
  \ifcase#1
    \or \textdagger       
    \or \textdaggerdbl    
    \or \textasteriskcentered 
    \or \S                
    \or \P                
    \else\@ctrerr
  \fi
}
\makeatother

\jmlrvolume{LEAVE UNSET}
\jmlryear{2025}
\jmlrsubmitted{LEAVE UNSET}
\jmlrpublished{LEAVE UNSET}
\jmlrworkshop{Conference on Health, Inference, and Learning (CHIL) 2025} 

\title[Time2Lang]{Time2Lang: Bridging Time-Series Foundation Models and Large Language Models for Health Sensing Beyond Prompting}

\author{
\Name{Arvind Pillai}$^1$ \Email{\normalsize arvind.pillai.gr{\small @dartmouth.edu}}  \\
\Name{Dimitris Spathis}$^{2,4,}$\thanks{This work is unrelated to the author's position at Google.}\Email{\normalsize spathis{\small @google.com}}  \\
\Name{Subigya Nepal}$^3$ \Email{\normalsize sknepal{\small @stanford.edu}}  \\
\Name{Amanda C Collins}$^1$ \Email{Amanda.C.Collins@dartmouth.edu}  \\
\Name{Daniel M Mackin}$^1$ \Email{Daniel.M.Mackin@dartmouth.edu}  \\
\Name{Michael V Heinz}$^1$ \Email{Michael.Vincent.Heinz@dartmouth.edu}  \\
\Name{Tess Z Griffin}$^1$ \Email{Tess.Z.Griffin@dartmouth.edu}  \\
\Name{Nicholas C Jacobson}$^1$ \Email{Nicholas.C.Jacobson@dartmouth.edu} \\
\Name{Andrew Campbell}$^1$ \Email{Andrew.T.Campbell@dartmouth.edu} \\
\addr $^1$Dartmouth College, USA \\
\addr $^2$University of Cambridge, UK \\
\addr $^3$Stanford University, USA \\
\addr $^4$Google Research, UK
}

\begin{document}

\maketitle

\begin{abstract}

Large language models (LLMs) show promise for health applications when combined with behavioral sensing data. Traditional approaches convert sensor data into text prompts, but this process is prone to errors, computationally expensive, and requires domain expertise. These challenges are particularly acute when processing extended time series data. While time series foundation models (TFMs) have recently emerged as powerful tools for learning representations from temporal data, bridging TFMs and LLMs remains challenging. Here, we present Time2Lang, a framework that directly maps TFM outputs to LLM representations without intermediate text conversion. Our approach first trains on synthetic data using periodicity prediction as a pretext task, followed by evaluation on mental health classification tasks. We validate Time2Lang on two longitudinal wearable and mobile sensing datasets: daily depression prediction using step count data (17,251 days from 256 participants) and flourishing classification based on conversation duration (46 participants over 10 weeks). Time2Lang maintains near constant inference times regardless of input length, unlike traditional prompting methods. The generated embeddings preserve essential time-series characteristics such as auto-correlation. Our results demonstrate that TFMs and LLMs can be effectively integrated while minimizing information loss and enabling performance transfer across these distinct modeling paradigms. To our knowledge, we are the first to integrate a TFM and an LLM for health, thus establishing a foundation for future research combining general-purpose large models for complex healthcare tasks.
 
\end{abstract}

\paragraph*{Data and Code Availability} In this work, we use the StudentLife dataset \citep{wang2014studentlife} which is publicly available. Another dataset, the Major Depressive Disorder (MDD) dataset, is currently private, with plans for public release in the future. Code pertaining to public data is available here\footnote{\url{https://github.com/arvind1609/time2lang}}.

\paragraph*{Institutional Review Board (IRB)} The study protocol for the major depressive disorder dataset has received full approval from the institutional review board (IRB) at Dartmouth College.

\section{Introduction}
Personal sensing devices such as mobile phones and wearables collect diverse behavioral data that can be used to assess mental health. Longitudinal data from these devices, including step counts and conversation patterns, effectively detect mental health conditions like depression \citep{nepal2024capturing, price2023predicting}. Meanwhile, large language models (LLMs) have recently demonstrated their ability to process a variety of data modalities beyond text, including audio \citep{ghosh2024gama} and images \citep{radford2021learning}. A notable strength of LLMs is their in-context learning capability, enabling them to generalize from prompt-based examples without additional training \citep{dong2022survey}. Building on these advancements, integrating behavioral sensing data with LLMs offers significant promise for developing personalized health models, with considerable impact for real-world applications \citep{cosentino2024towards}.


\begin{figure}
    \centering
    \includegraphics[width=1\linewidth]{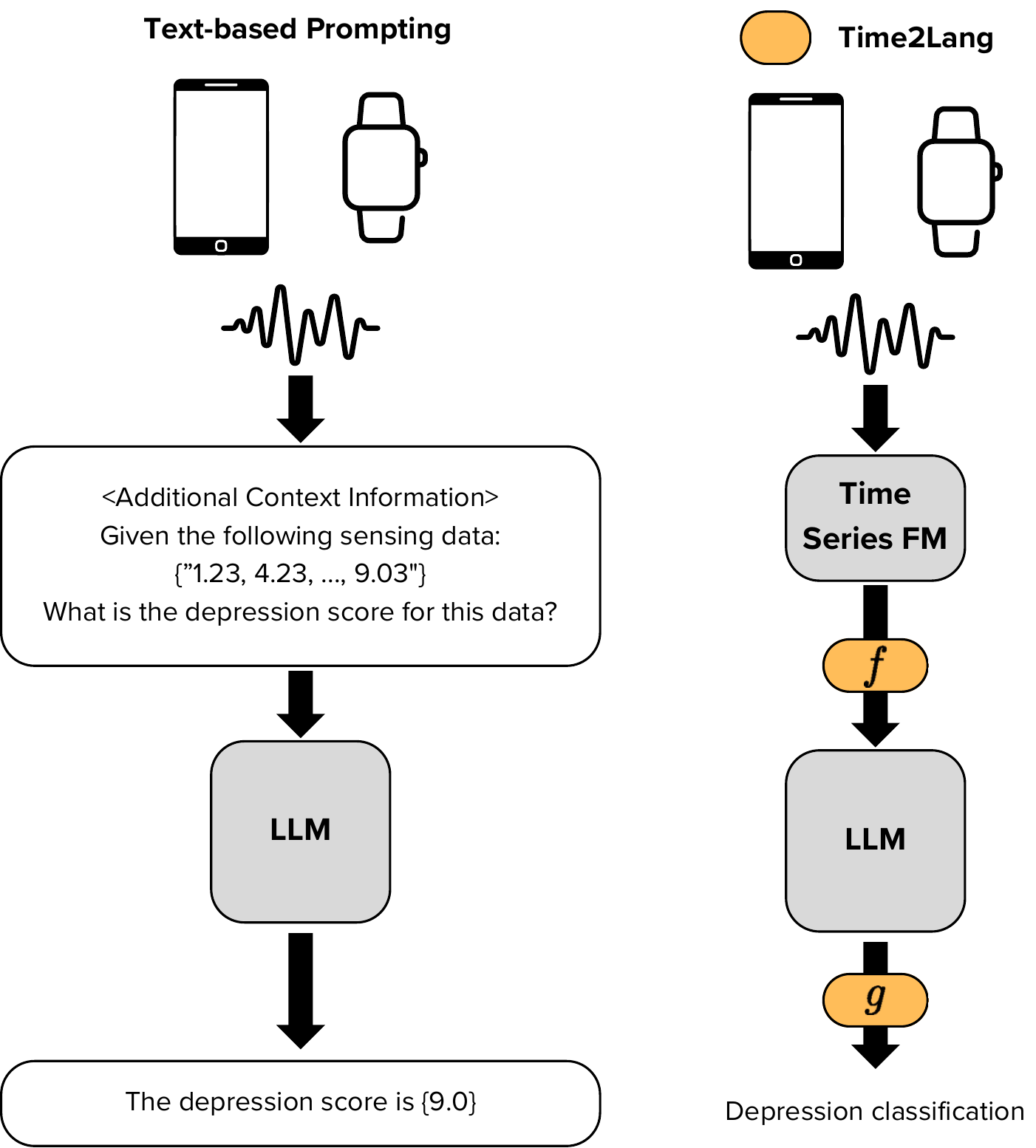}
    \caption{\textbf{Time2Lang vs traditional prompting.} An example of using sensing data to predict depression, comparing text-based prompting (left) with Time2Lang (right). In text-based prompting, the sensor signals are converted into text for LLM prompting. As an alternative, we introduce Time2Lang—our method learns a mapping ($f$ and $g$) between a TFM and an LLM, enabling health sensing tasks without the need for text conversion, while making the most of powerful models.}
    \label{fig:prompting_Time2Lang}
\end{figure}

Recent studies have incorporated health sensing data with LLMs to detect depression \citep{kim2024health}, atrial fibrillation \citep{liu2023large}, and sleep quality \citep{cosentino2024towards}. A common approach involves converting raw sensor data into text prompts to enable LLMs to process numeric time series data for classifying these conditions \citep{kim2024health, liu2023large}. However, this method ignores the modality gap between sensing and text data \citep{spathis2024first}. Applying this technique to long-duration time series leads to higher error rates during prompting and requires a large amount of tokens (Section \ref{sec:limitations}). Constructing effective prompts also requires substantial domain expertise, making it a challenging task even for domain experts (Section \ref{sec:baselines}). 
Recent advancements in time series foundation models (TFM), such as Chronos \citep{ansari2024chronos} and Moment \citep{goswami2024moment}, have demonstrated their effectiveness in generating meaningful representations by capturing fundamental time series characteristics. 

Therefore, we explore if incorporating a TFM in an LLM pipeline can address the aforementioned limitations. One way to efficiently incorporate two large models from different domains is through the concept of \textit{model re-programming} \citep{chen2024model}. Previous works have successfully reprogrammed models from distinct domains such as text to biochemical sequence \citep{vinod2023reprogramming} and speech to time series \citep{yang2021voice2series}. Hence, we argue that incorporating a TFM into an LLM model through reprogramming can produce meaningful representations. Specifically, rather than using raw signals in prompts, we leverage TFMs as feature extractors (Figure \ref{fig:prompting_Time2Lang}). A key question is whether an LLM can understand TFM-derived features. To this end, we introduce an adapter that maps TFM features to LLM representations, effectively serving as a translator between the two models (Figure \ref{fig:prompting_Time2Lang}). We demonstrate our approach through behavioral sensing applications in mental health. Since behavioral sensing data is collected longitudinally in real-world settings, its unobtrusive nature results in long-duration sequences with realistic noise. Moreover, mental health is a high-stakes domain where minimizing errors, enhancing time-series comprehension, and improving efficiency are critical. Rather than proposing a new model, our goal is to maximize the potential of existing powerful models through an elegant adapter strategy. Thus, we explore whether two large models from distinct domains can be integrated with minimal information loss.

Toward the vision of seamless integration of time series data into LLMs, our contributions are as follows: (1) \textbf{Time2Lang Framework.} We propose Time2Lang, a framework that learns a mapping between two frozen foundation models (FMs): Chronos \citep{ansari2024chronos} and LLaMA \citep{dubey2024llama} (Figures \ref{fig:prompting_Time2Lang} and \ref{fig:Time2Lang}). To our knowledge, we are the first to integrate a TFM and an LLM for health tasks. Unlike most prior work, Time2Lang is trained within a self-supervised framework exclusively using intricate synthetic data generated from a Gaussian process (Section \ref{sec:synthetic}). (2) \textbf{Evaluation on Longitudinal Mental Health Data.} We assess Time2Lang’s performance on two real-world longitudinal mental health datasets with long-duration sequences. Specifically, we predict daily depression levels using step count data (17,251 days/labels) and analyze student flourishing from conversation duration (46 participants over 10 weeks) as described in Section \ref{sec:experiments}. (3) \textbf{Efficiency and Temporal Dynamics.} We examine Time2Lang’s inference time, the impact of pre-training data size, the role of periodicity, and its relationship to classical time-series characteristics (Section \ref{sec:results}). Our \textbf{key findings} (Section \ref{sec:findings}) are: (1) Time2Lang’s learned mapping minimizes information loss and enables positive transfer between two distinct domains—time series and text, (2) it maintains near constant inference time per sample with increasing input lengths, and (3) its mapped LLM outputs align with fundamental time-series metrics, such as the autocorrelation factor. Overall, Time2Lang demonstrates that reprogramming can effectively integrate a TFM with an LLM, enhancing performance, efficiency, and time-series comprehension.



\section{Related Work}

\subsection{Model Reprogramming}

Model reprogramming is a training paradigm that efficiently repurposes a pre-trained, frozen large model across domains using an input transformation layer and an output mapping layer \citep{chen2024model}. \citet{yang2021voice2series} reprogrammed a speech model for univariate time series classification, while \citet{neekhara2022cross} proposed an adversarial approach to convert an image model for protein sequence prediction. Similarly, \citet{vinod2023reprogramming} focused on mapping a text model to biochemical sequences. From a reprogramming perspective, our work introduces several key advancements. It demonstrates that a TFM and an LLM can be effectively integrated for health-related tasks. Unlike most prior work, we adopt a self-supervised learning (SSL) framework instead of supervised learning. Additionally, our approach explicitly incorporates periodicity properties, further enhancing time-series understanding.



\subsection{LLMs and Healthcare}
Using LLMs to evaluate health outcomes has seen rapid growth \citep{he2023survey, han2023medalpaca, liu2025generalist, singhal2025toward}. For instance, MedPaLM 2 \citep{singhal2025toward}, an LLM fine-tuned for medical question answering has demonstrated strong performance by passing medical exam benchmarks. Similarly, models such as MedAlpaca \citep{han2023medalpaca} and Me-LLaMA \citep{xie2024me} also fine-tune general-purpose LLMs for comprehensive analysis of medical texts. For mental health tasks, \citet{xu2024mental} comprehensively evaluated different LLMs using online text data. Their work demonstrated that incorporating additional health context information in prompts enhances performance, but specialized domain knowledge is required.

\begin{figure}
    \centering
    \subfigure[]{\includegraphics[width=0.24\textwidth]{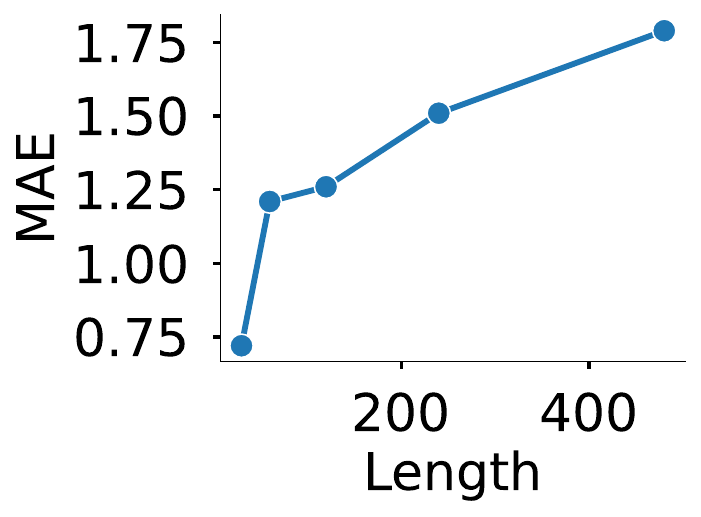}} 
    \subfigure[]{\includegraphics[width=0.24\textwidth]{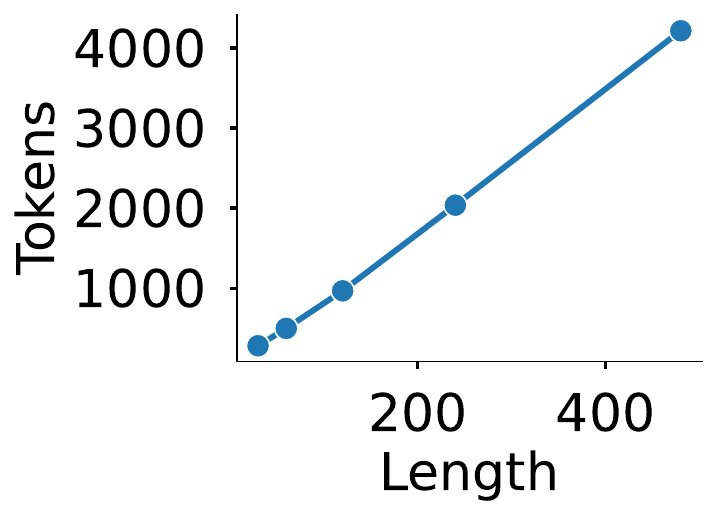}} 
    \caption{\textbf{Effect of Increasing Sequence Length on Prompting Performance.} We evaluate LLaMA 3.2 (1B) on: (a) a mean prediction task, where mean absolute error increases with sequence length, and (b) tokens, where the number of tokens is $\sim 10\times$ the time-series length.} 
    \label{fig:limitations}
\end{figure}

\subsection{LLMs and Sensor Data}

Health sensing data consists of temporal numeric values that change over time. Approaches for time-series forecasting with LLMs involved converting the raw data into text prompts, as demonstrated by methods like LLMTime \citep{gruver2024large} and PromptCast \citep{xue2023promptcast}. For instance, \citet{gruver2024large} highlighted the importance of careful pre-processing, such as handling periods (“.”) and spaces (“ ”), to ensure effective tokenization of floating-point numbers.  Recently, alternative approaches have used multi-modal LLMs that encode sensing data as images \cite{yoon2024my} or use modality-specific encoders \cite{belyaeva2023multimodal}. For example, \citet{yoon2024my} proposed transforming time-series data into visual prompts for use with vision-language models.

\begin{figure*}
    \centering
    \includegraphics[width=1\linewidth]{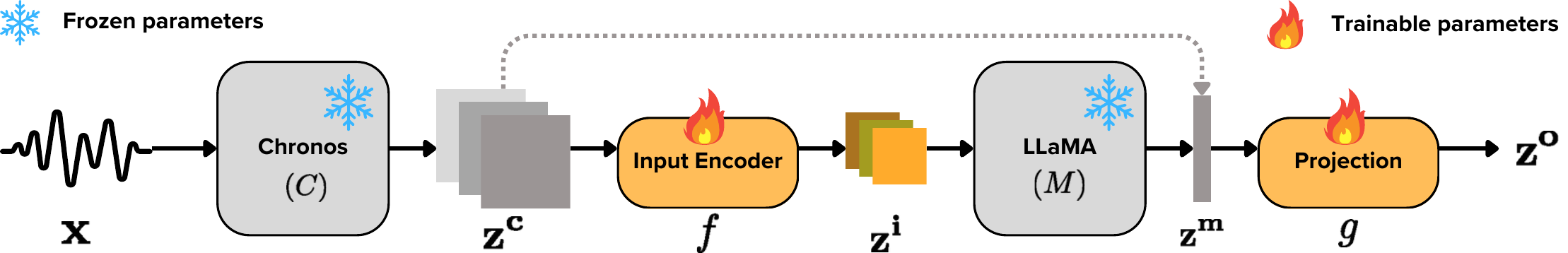}
    \caption{\textbf{Time2Lang Framework.} To meaningfully integrate Timeseries Foundation Models  (here: Chronos $C$) and Large Language Models (LLaMA $M$), we train two smaller networks $f$ and $g$ that optimally map TFM features ($\mathbf{z^c}$) to an LLM. The learned embeddings from $f$ and $g$ are $\mathbf{z^i}$ and $\mathbf{z^o}$, respectively. To improve positive knowledge transfer, we use a residual connection between the TFM and LLM features ($\mathbf{z^c} \rightarrow \mathbf{z^m}$) only during training.}
    \label{fig:Time2Lang}
\end{figure*}

\subsection{Limitations of integrating mobile sensing with LLMs} \label{sec:limitations}
Despite advances for integrating sensor data into LLMs, the most common approach remains incorporating the raw sensor data as text strings into prompts without extensive pre-processing \citep{kim2024health, liu2023large, xu2024penetrative}. While this technique has shown promise for aggregated short-duration time series (e.g., $<$ 200 time steps), behavioral sensing data often has long duration sequences. For example, one day of data in one minute intervals consists of 1440 data points. Sampling rates $>1$Hz will have substantially more data points. Therefore, representing sensor data as text has the following limitations. First, an LLM performance decreases with longer duration sequences \citep{yoon2024my}. To empirically demonstrate this using LLaMA 3.2 \citep{dubey2024llama}, we perform a mean prediction task using synthetic data, where LLaMA is asked to predict the mean with increasing sequence length. In Figure \ref{fig:limitations}, we notice that the mean absolute error (MAE) increases with increasing sequence length. Second, long duration sequences require significantly more tokens, which may quickly exhaust the models limits. Specifically, observe that the number of tokens required for LLaMA 3.2 is $\sim 10\times$ the length of the time series (Figure \ref{fig:limitations}). Third, effective prompting is challenging and requires strong domain knowledge (adding specific knowledge improves prompting performance -- Table \ref{tab:overall_performance} LLaMA-ICL vs LLaMA-ICL-DK), remaining an active research area in NLP \citep{zamfirescu2023johnny, bommasani2021opportunities, liu2023pre}.

Our approach offers an alternative perspective for addressing these limitations with mental health as an application. By learning a mapping that seamlessly integrates a TFM into the LLM workflow, \textit{Time2Lang inputs fixed-size embeddings directly into the LLM, eliminating the need for prompting}. We show that the Time2Lang framework is more scalable and efficient compared to generative LLM prompting, while preserving the performance of the TFM.

\section{Methods} \label{sec:methods}

\subsection{Synthetic Data Generation} \label{sec:synthetic}
Human behavioral signals often display periodic patterns, and leveraging self-supervised learning (SSL) to capture these patterns has been shown to produce robust representations \citep{yang2022simper, gu2019home}. To enable Time2Lang to learn periodicity, we generate synthetic data that mimics intricate sensing patterns by sampling from a Gaussian Process (GP) regression. In particular, we modify KernelSynth \citep{ansari2024chronos} to produce time series that exactly repeat themselves after a specific interval. A GP is defined by a mean function $m(a)$ (set to 0) and a covariance function or kernel $k(a, a')$ which models the joint variability of the pair of points $(a, a')$ \citep{schulz2018tutorial}. 

We perform the following steps to generate the synthetic data. First, we randomly sample up to $j=4$ non-periodic kernels from a bank $\mathcal{K}$ consisting of the following kernels: constant, white noise, linear, radial basis function, and rational quadratic ($\{k_1(a, a'), \cdots, k_j(a, a')\}$). These kernels are useful to model general sensor patterns such as trend, noise, and local variation. Second, we sample a periodic value $p$ from a set of pre-defined values $p \in \{30, 60, 90, 120, 150, 180\}$, and generate the exponential sine squared kernel, $k_{per}(a, a')$. This infuses seasonality into the generated data such that the patterns exactly repeat themselves. Third, given a set of non-periodic and periodic kernels $\{k_1(a, a'), \cdots, k_j(a, a'), k_{per}(a, a')\}$, we compose them into a single $k(a, a')$ by applying a random binary operation: $+$ or $\times$. Finally, a synthetic time series of length $l=1440$ (one day of data in one-minute intervals) is generated by sampling from the GP with $m(a)=0$ and covariance function $k(a, a')$. 

\subsection{Time2Lang} \label{sec:Time2Lang}

Time2Lang consists of four sequential components (Figure \ref{fig:Time2Lang}), a frozen TFM ($C$), a learnable input encoder ($f$), a frozen LLM ($M$), and a learnable projection ($g$). It is worth noting that inputs to the input encoder and the projection are the outputs of $C$ and $M$, respectively. 

\subsubsection{Training} 
We train Time2Lang within an SSL framework using synthetic data, where periodicity prediction is the pretext task. Formally, given a time series $\mathbf{x} \in \mathbb{R}^{n}$ of length $n$ and the corresponding periodicity $y \in \{30, 60, 90, 120, 150, 180\}$, the objective is to learn $f$ and $g$ to accurately predict periodicity as a multi-class classification problem.

Time2Lang is trained following the steps outlined in Algorithm \ref{alg:training}. Given an pair $(\mathbf{x}, y)$, we first extract features from $\mathbf{x}$ using a frozen TFM - Chronos ($C$) \citep{ansari2024chronos}. The resulting Chronos embeddings $\mathbf{z^c} \in \mathbb{R}^{c \times d}$ are temporal features that are fixed irrespective of input size, where context length $c=513$ and features $d=768$. Previous work has shown that TFMs, such as Chronos, effectively extract meaningful time-series features. Instead of using raw signals as prompts, we leverage these extracted features. However, since there is no guarantee that an LLM can interpret these features, we aim to explicitly establish a meaningful mapping between TFM-derived features and an LLM, specifically LLaMA 3.2 ($M$). To achieve this, we train an input encoder ($f$), a lightweight 1D ResNet-style CNN. This encoder processes the transposed output of $C$, $\mathbf{z^c}^T$, generating compressed temporal features $\mathbf{z^i} \in {\mathbb{R}^{d \times c}}$, where the feature dimension $d$ and sequence length $c$ are determined by the convolution and pooling parameters ($d=65$, $c=64$ in our case). We apply zero padding to $\mathbf{z^i}$ up to a length of 2048, matching the input embedding size of LLaMA. The encoded representation, $\mathbf{z^i}^T$, is then directly passed into LLaMA's ($M$) input embedding layer instead of using token ids, thus by-passing the tokenization process. The output is averaged to obtain $\mathbf{z^m} \in \mathbb{R}^{2048}$. Subsequently, we project $\mathbf{z^m}$ using two fully-connected (FC) layers with 768 and 256 neurons. Additionally, a residual connection is introduced between the averaged $\mathbf{z^c}$ and the first FC layer to enhance information retention from Chronos. Finally, $\mathbf{z^o}$ is passed through a output FC layer $l$ to produce a logit vector $\mathbf{\hat{y}}$. Conceptually, the input encoder ($f$) and projection network ($g$) are trained to optimize the probing of the frozen LLaMA model, leveraging periodicity as a pretext task to learn useful embeddings. The model is trained end-to-end by optimizing the multi-class cross entropy loss as follows:

\begin{equation} \label{eqn:cross_entropy}
\mathcal{L} = -\frac{1}{N}\sum_{k=1}^{N} \log \frac{\exp{(\mathbf{\hat{y}}_{k,y_k})}}{\sum_{j=1}^{6}\exp{(\mathbf{\hat{y}}_{k,j})}}
\end{equation}
where $N$ is batch size, $y_k$ is the correct index for sample $k$, and $6$ represents the number of periodic classes.


\begin{algorithm2e}[t]
\caption{Time2Lang Training} \label{alg:training}
\algorithmfootnote{\textsuperscript{\textdagger} mean is computed over the context length dimension}
\KwIn{$N$ synthetic time series with periodicity $\{\mathbf{x}_k, y_k\}_{k=1}^{N}$, TFM: $C$, LLM: $M$, Input Encoder: $f$, Projection: $g$, Output layer: $l$.}
\For{$k \in \{1, \dots, N\}$}{
     $\mathbf{z^c}_k \leftarrow C(\mathbf{x}_k)$ \\
     $\mathbf{z^i}_k \leftarrow f((\mathbf{z^c}_k)^T)$ \\
     $\mathbf{z^m}_k \leftarrow M((\mathbf{z^i}_k)^T)$ \\
     $\mathbf{z^m}_k \leftarrow \text{mean}(\mathbf{z^m}_k)$ \\
     $\mathbf{z^o}_k \leftarrow g(\mathbf{z^m}_k, \text{mean($\mathbf{z^c}_k$))}$ \# residual\\
     $\mathbf{\hat{y}}_k \leftarrow l(\mathbf{z^o}_k)$
} 
Compute $\mathcal{L} = -\frac{1}{N}\sum_{k=1}^{N} \log \frac{\exp{(\mathbf{\hat{y}}_{k,y_k})}}{\sum_{j=1}^{6}\exp{(\mathbf{\hat{y}}_{k,j})}}$ \\
Update networks $f$ and $g$ to minimize $\mathcal{L}$.
\end{algorithm2e}

Our rationale for choosing Chronos and LLaMA is based on several key considerations. First, Chronos draws inspiration from language modeling, specifically, its approach to time series quantization and discretization is analogous to tokenization in language models. Second, it transforms variable-length inputs into fixed-size embeddings, enhancing efficiency as input lengths grow. Lastly, these models are well-established within their respective fields, promoting the development of findings that are more broadly generalizable.

\subsubsection{Implementation} \label{sec:implementation}
We generate 200K synthetic samples, as outlined in Section \ref{sec:synthetic}. These samples are further split into training, validation, and testing set in 70/10/20 ratio. Time2Lang is trained \textit{exclusively} on synthetic samples for 25 epochs, using a batch size of 16, on an NVIDIA A4500 GPU. The training process optimizes the model to predict periodicity by minimizing the cross-entropy loss (Equation \ref{eqn:cross_entropy}). Specifically, the Adam optimizer with a learning rate of $5 \times 10^{-4}$ is used to update the parameters of the networks $f$ and $g$. Given the sensitive nature of mental health data, we implement the model locally, utilizing Chronos base and LLaMA 3.2 as frozen foundation models (FMs) with 200M and 1B parameters, respectively. Moreover, if our approach works for these smaller models, it will improve with larger models due to scaling laws \citep{ansari2024chronos, kaplan2020scaling}. The input encoder $f$ consist of 300K parameters ($\approx 0.03\%$ of LLaMA), while the projection $g$ consists of 1.7M parameters ($\approx0.17\%$ of LLaMA).

\section{Experimental Setup} \label{sec:experiments}
In this section, we describe our rationale for choosing datasets, baselines, and evaluation protocol.

\subsection{Datasets} \label{sec:datasets}
We sought to evaluate Time2Lang on diverse longitudinal datasets in terms of sensing modality, devices, population, and tasks. Therefore, we examine: (1) daily depression detection in a clinical population using step count data from a wearable device, and (2) classification of flourishing levels in a student population (non-clinical) based on their conversation duration from mobile phones over the course of an academic term. We describe dataset information relevant to our analyses below. 

\subsubsection{Major Depressive Disorder (MDD)}
The Major Depressive Disorder (MDD) longitudinal study investigates intra-day fluctuations in depressive symptoms among individuals diagnosed with MDD. In this 90-day study, we recruited participants aged 18 years and older, residing in the United States. To determine eligibility, we administered the structured clinical interview for DSM-5 (SCID), ensuring that only individuals with MDD, but without co-morbid bipolar disorder, active suicidality, or psychosis, were enrolled. Each participant was instrumented with a Garmin vivoactive 3 wearable and installed our Android application. While the Garmin passively collected sensing data, the mobile application prompted the user to answer the Patient Health Questionnaire-9 (PHQ-9) three times a day \citep{kroenke2001phq}. This self-reported survey consisting of nine questions rated on a 0-3 Likert scale is a well-validated tool to evaluate depression. To encourage engagement, participants received \$1 per completed survey, with an additional \$50 bonus for maintaining a survey completion rate greater than 90\%. Now, we describe information pertaining to the analysis in this work. Participant safety and ethics are discussed in Appendix~\S\ref{apd:ethics}.

Physical activity plays a crucial role in depression fluctuations, and step count provides a straightforward measure of activity levels. Notably, a daily step count exceeding 5,000 has been associated with fewer depressive symptoms \citep{bizzozero2024daily}. Hence, we assess daily depression levels using step count data from the Garmin wearable. It is worth noting that daily depression prediction is challenging problem compared to longer time-scales. In our analysis, the input sensing data consists of 1,440 minute-level step count values per day, while the ground truth is derived from the Patient Health Questionnaire-9 (PHQ-9), where a score of 10 or higher indicates depression (Figure \ref{fig:phq9}). To handle missing data, we apply a threshold-based approach: if missingness is below 25\%, we impute missing values using zero-filling; otherwise, we exclude the day from the analysis. The PHQ-9 score is averaged within each day to align with the step count data. Our dataset comprises 256 participants, covering a total of 17,251 person-days (68\% depressed days), with corresponding depression labels.

\subsubsection{StudentLife}
The StudentLife study \citep{wang2014studentlife} tracked the behavior of 48 undergraduate students at Dartmouth College over a 10-week term using mobile sensing. An Android application passively collected conversation duration from ambient sound, while self-reported surveys were administered to examine students' mental health. Beyond academics, psychological well-being is an important part of college life, involving personal growth, social connections, and civic engagement. The Flourishing Scale (FS) \citep{diener2010new} measures this well-being through an 8-item questionnaire rated on a 1-7 scale from strongly disagree to strongly agree (Appendix~\S\ref{apd:emas}). The total FS score provides a single measurement of psychological well-being. As social contact is directly examined in FS, we focus on classifying students into high and low flourishing (median split) groups based on their conversation duration throughout the academic term. Specifically, the input is a variable-length conversation duration time series up to 10 weeks, where each data point represents hourly conversation duration, and the output is a binary indicator of high or low flourishing. We use data from 46 students with an average time series length of $1455 \pm 193$. 

\begin{table*}[] 
\centering
\caption{\textbf{Time2Lang performance comparison against baselines}. Values in \textbf{bold} denote when models outperform the upper bound (Chronos). Higher values indicate better performance.}
\begin{tabular}{lllll}
\toprule
\multirow{2}{*}{\textbf{Model}} & \multicolumn{2}{c}{\textbf{Depression (step count)}} & \multicolumn{2}{c}{\textbf{Flourishing (conversation)}} \\ \cmidrule{2-5}
    & AUROC                 & AUPRC                & AUROC                  & AUPRC                 \\ \midrule
 Random (Lower) &0.50 &0.68 &0.50 &0.52 \\
 Chronos (Upper) &0.56 (0.01) &0.72 (0.01) &0.70 (0.15) & 0.69 (0.14) \\
 \midrule
 LLaMA-ICL &0.49 (0.03) &0.67 (0.02) &0.51 (0.11) &0.61 (0.09) \\
 LLaMA-ICL-DK &0.53 (0.03) & 0.68 (0.02) &0.54 (0.10) & 0.63 (0.15) \\
 Chronos + LLaMA  &0.54 (0.01) &0.70 (0.00) &0.56 (0.16) & 0.65 (0.15) \\
 Time2Lang &\textbf{0.57} (0.01) &\textbf{0.73} (0.00) & \textbf{0.71} (0.10) &\textbf{0.74} (0.11) \\
 \midrule
 Time2Lang-L (Section \ref{sec:architecure}) &0.57 (0.00) &0.73 (0.00) &0.80 (0.06) &0.85 (0.05) \\
 \bottomrule
 \label{tab:overall_performance}
\end{tabular}
\end{table*}

\subsection{Baselines} \label{sec:baselines}
We evaluate Time2Lang using several baseline approaches. The lower bound baseline is \textbf{Random} classification where AUROC is 0.5 and AUPRC is the proportion of positive samples. As we transfer information from Chronos to LLaMA, we treat \textbf{Chronos} as the upper bound baseline, assuming some information loss occurs during translation. To determine whether reprogramming through a mapping function is necessary, we directly embed Chronos into the LLaMA embedding without a mapping function (with embeddings padded as needed), referred to as \textbf{Chronos + LLaMA}. To compare against prior prompting methods, we use a 3-shot in-context learning (ICL) approach with LLaMA, referred to as \textbf{LLaMA-ICL}. The prompt design follows established best practices from existing literature. First, from LLMTime \citep{gruver2024large}, we round numerical values to two-digit precision, remove periods without adding spaces, and enclose the time series values in quotes to convert them into strings (e.g., $0.123 \rightarrow ``012"$). Second, building on HealthLLM \citep{kim2024health}, we include health-related context (e.g., ``Some symptoms of depression include a lack of motivation and fatigue, which can be inferred from daily step counts.") and temporal context (e.g., ``\{20, 25, ..., 30"\}). To further support LLaMA-ICL, we incorporate domain-specific knowledge -- \textbf{LLaMA-ICL-DK}, which is crucial for guiding the model effectively. For example, ``Conversation duration is associated with flourishing, with longer durations typically indicating higher levels of flourishing." This addition from the self-reported instrument provides more context to better understand and interpret the data within the specific application domain.

\subsection{Evaluation} \label{sec:evaluation}
After initial pre-training, Time2Lang can be used for a wide variety of different time-series tasks without further fine-tuning. For downstream evaluation, we process the input sensing data from the real datasets and extract embeddings from the penultimate linear layer \textit{without} the residual connection to determine whether the information from Chronos has been effectively distilled. The extracted embeddings are then used to train either a logistic regression (LR) or random forest (RF) model for mental health classification. Prior to training, we split the dataset into an 80/20 subject-wise split for training and hold-out testing. Next, we perform grid search cross-validation (3 folds) on the training data to identify the optimal model parameters. Once the best parameters are selected, we evaluate the final model on the hold-out test set using the area under the receiver operating characteristic curve (AUROC) and the area under the precision-recall curve (AUPRC). AUROC evaluates the overall discriminative power of the model, while AUPRC places greater emphasis on the performance for the positive class. Moreover, we repeat the experiments across five different shuffled training splits and report the standard deviations of the results for the best model (LR or RF). 

\section{Results} \label{sec:results}

\subsection{Overall Performance}

Table \ref{tab:overall_performance} presents the performance comparison of Time2Lang with the different baselines for depression detection and flourishing classification. For depression detection using step count, Time2Lang achieves an AUROC of 0.57 and an AUPRC of 0.73, outperforming the baselines. Compared to Chronos, Time2Lang slightly improves the AUROC (0.56 vs. 0.57) and AUPRC (0.72 vs. 0.73) for depression detection. For flourishing classification using conversation duration, Time2Lang surpasses the prompting-based and Chronos + LLaMA baselines, with AUROC and AUPRC improvements ranging from 0.15 to 0.20 and 0.09 to 0.13, respectively. Compared to Chronos, Time2Lang achieves an AUROC of 0.71 (+0.01) and an AUPRC of 0.74 (+0.05). The higher AUPRC suggests that Time2Lang improves positive case detection. Overall, Time2Lang consistently outperforms most baselines, demonstrating marginal gains over Chronos in three cases and a more substantial improvement in one case.

Some other interesting observations from Table \ref{tab:overall_performance} are as follows.  First, incorporating domain knowledge into prompting enhances performance across tasks in both AUROC and AUPRC (LLaMA-ICL vs. LLaMA-ICL-DK). Notably, for depression detection, LLaMA-ICL underperforms the lower-bound baseline, with an AUROC of 0.49 and an AUPRC of 0.67, suggesting that prompting for long-duration sequences poses challenges. However, adding domain knowledge improves AUROC by 0.04 and AUPRC by 0.01, demonstrating its positive impact. Second, integrating Chronos with LLaMA leads to a decline in performance, particularly in the flourishing prediction task, where AUROC and AUPRC drop by 0.14 and 0.04, respectively, compared to Chronos alone. In contrast, Time2Lang enables positive performance transfer, a 0.01 increase in AUROC while achieving a notable 0.05 increase in AUPRC, highlighting the necessity of having a mapping to bridge two distinct domains.

\subsection{Efficiency Analysis}

\begin{figure}
    \centering
    \subfigure[]{\includegraphics[width=0.24\textwidth]{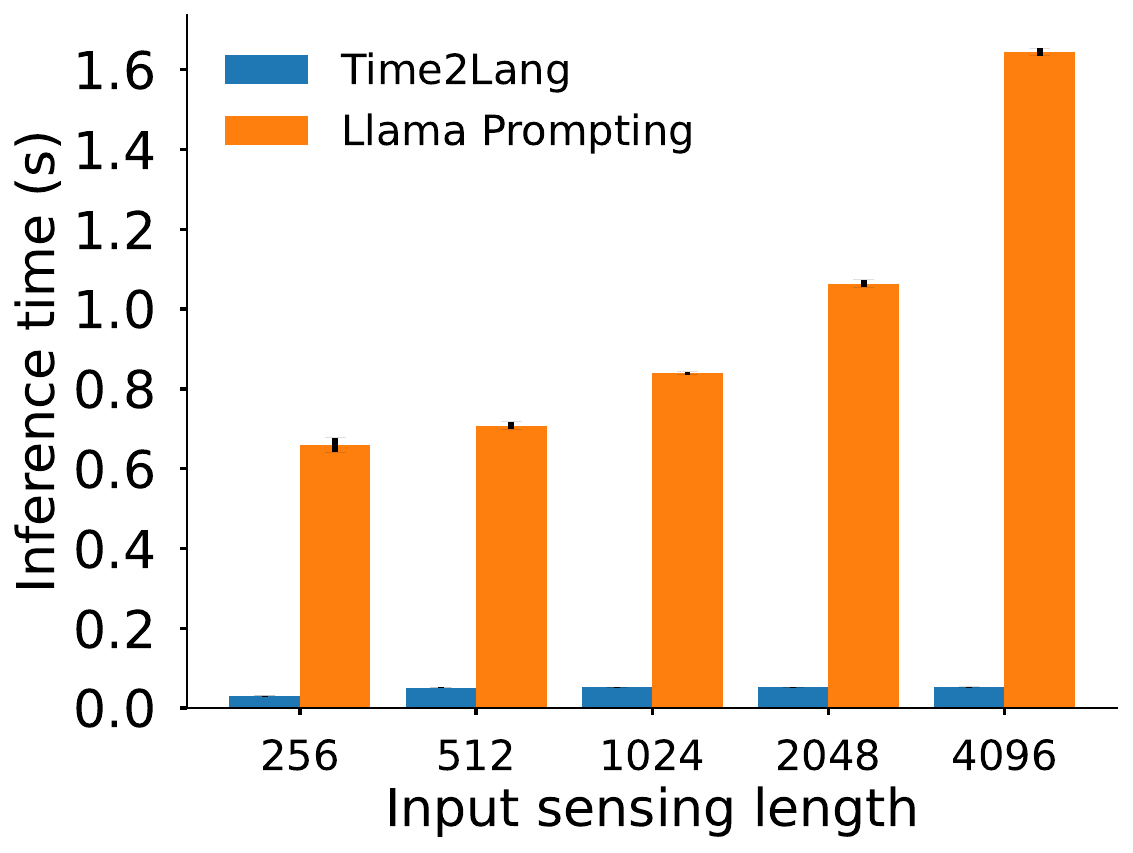}} 
    \subfigure[]{\includegraphics[width=0.24\textwidth]{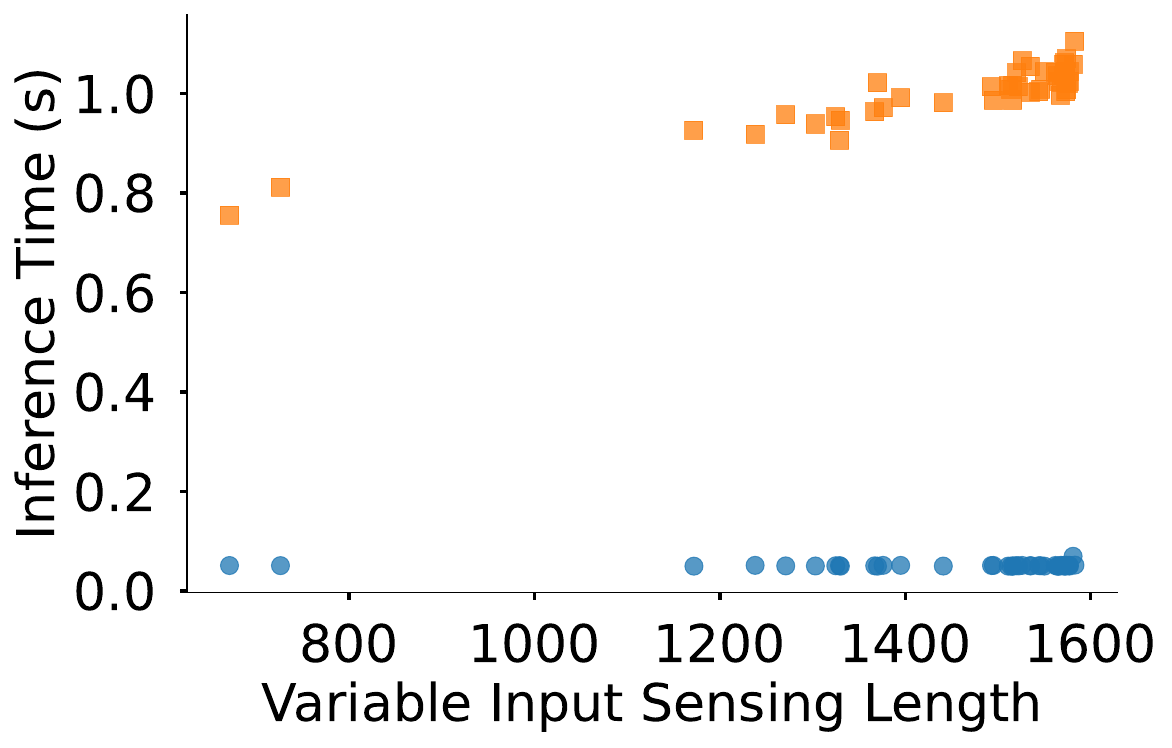}} 
    \caption{\textbf{Efficiency Analysis}. Inference time (seconds) per sample or Latency comparison between Time2Lang and Prompting for (a) different input sequence lengths, (b) variable-length conversation duration data from StudentLife. We repeat measurements 100 times and observed a variance of $<10^{-2}$.}
    \label{fig:efficient}
\end{figure}

To evaluate the efficiency of Time2Lang and LLaMA prompting, we assess inference times using two approaches. First, we measure inference time across varying time series lengths: $\{256, 512, 1024, 2048, 4096\}$ (Figure \ref{fig:efficient}(a)). Second, given that the StudentLife conversation data represents a variable-length time series, we evaluate inference time across all data samples (Figure \ref{fig:efficient}(b)). Prior to measurement, we ensure consistent execution by transferring the data and model to the device, synchronizing GPU, and performing a warm-up phase with 100 iterations. As shown in Figure \ref{fig:efficient}, Time2Lang consistently achieves lower inference times compared to prompting across different data lengths in the synthetic experiment and conversation duration data in the StudentLife dataset. Notably, the inference time for Time2Lang remains relatively stable as the data length increases, whereas LLaMA prompting exhibits a proportional increase in inference time. This disparity is primarily due to the increasing relationship between input token length and sensing data length, which affects the computational overhead of prompting-based approaches. Numerical results demonstrating Time2Lang's superior latency and throughput compared to LLaMA prompting are described in Appendix~\S\ref{apd:efficiency_analysis}. Notably, while LLaMA prompting experiences a decline in efficiency with increasing input lengths, Time2Lang maintains near-constant latency and throughput across input sizes ranging from 512 to 4096, highlighting its scalability and efficiency.

\subsection{Ablation Study: Pre-training Data and Periodicity Classes}

\begin{figure}
    \centering
    \subfigure[]{\includegraphics[width=\linewidth]{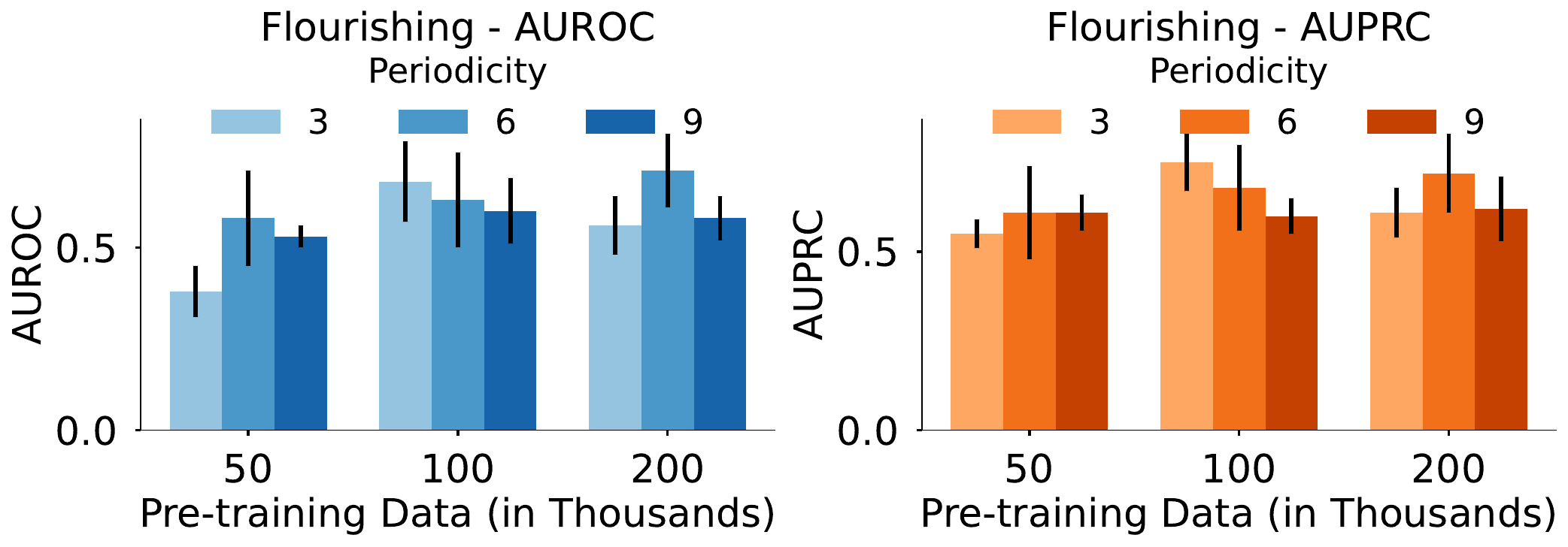}} 
    \subfigure[]{\includegraphics[width=0.5\textwidth]{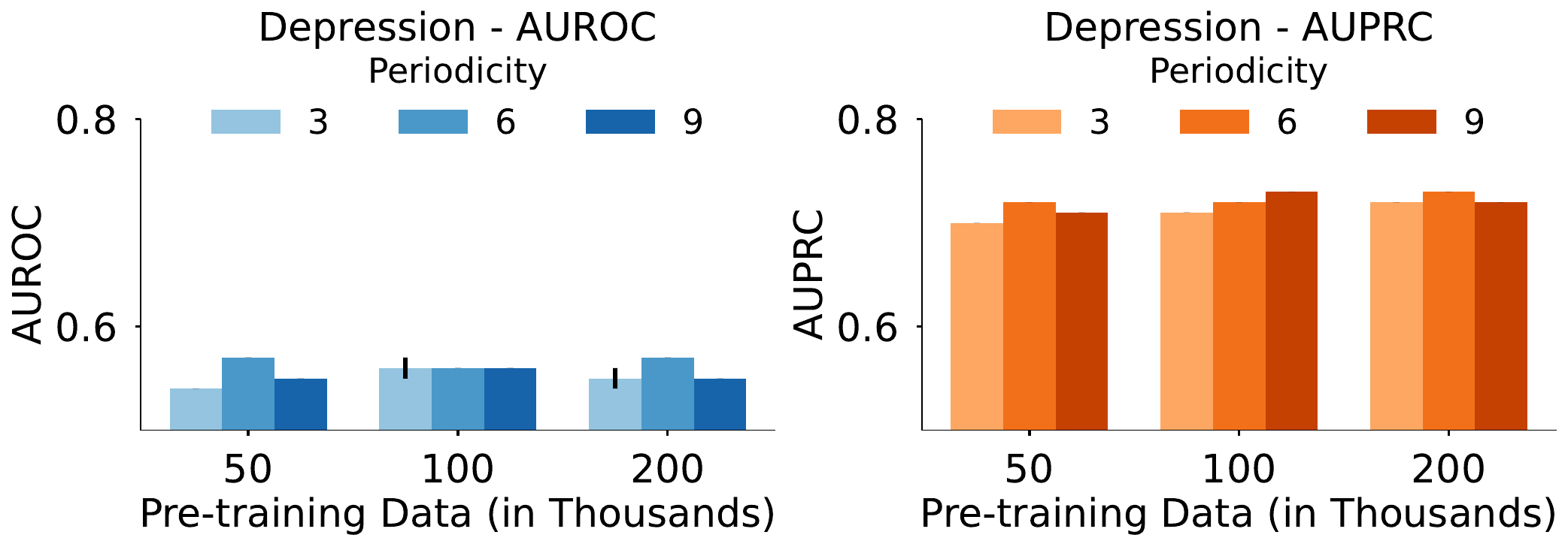}} 
    \caption{\textbf{Pre-training ablations}. The effect of pre-training data size and periodicity classes on downstream performance. Optimal performance is achieved with 200K samples in a 6-class classification problem.}
    \label{fig:effect_of_pre_training}
\end{figure}

\begin{figure*}
    \centering
    \subfigure[]{\includegraphics[width=0.34\linewidth]{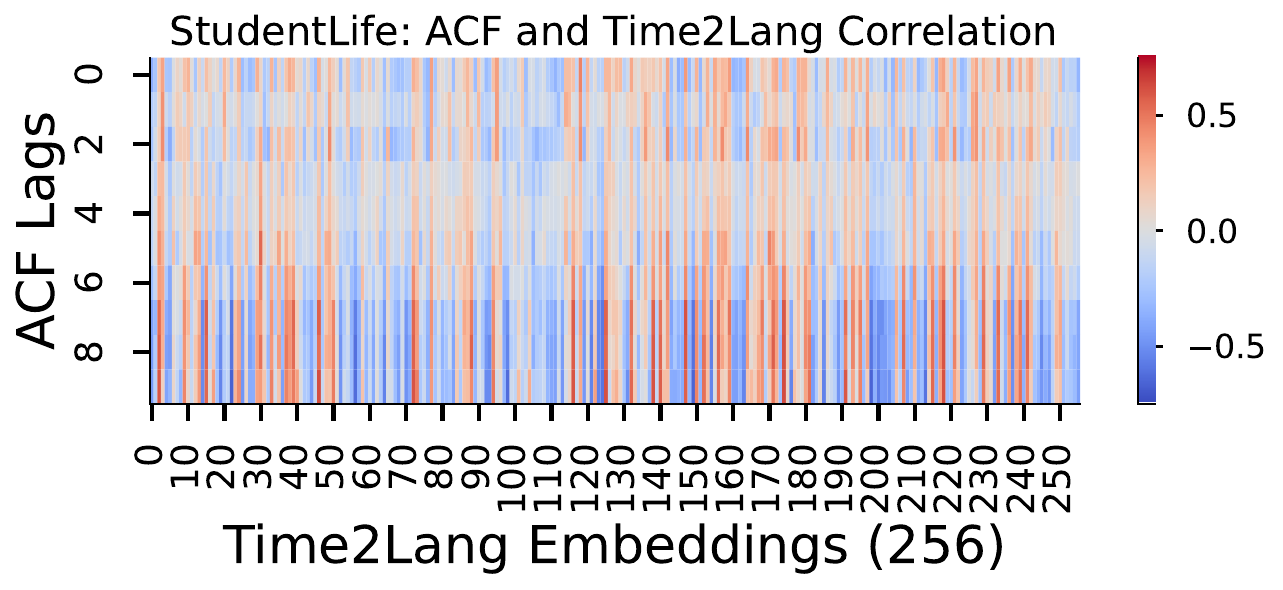}}
    \subfigure[]{\includegraphics[width=0.15\linewidth]{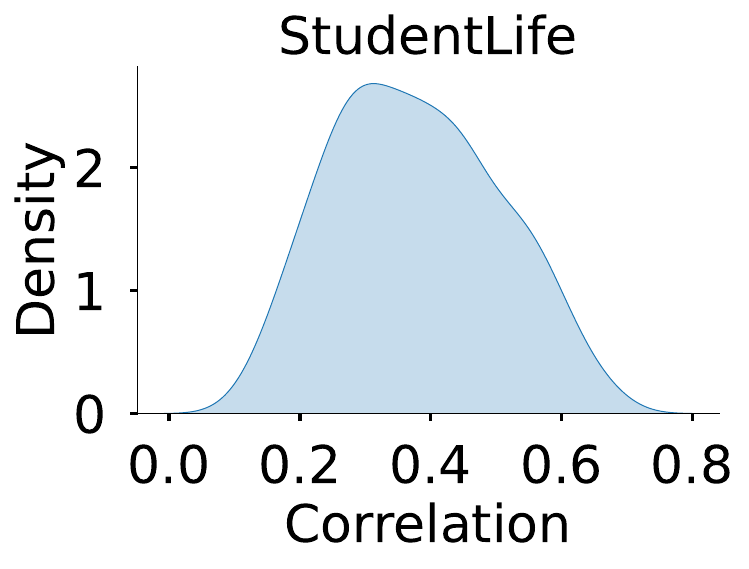}}
    \subfigure[]{\includegraphics[width=0.34\linewidth]{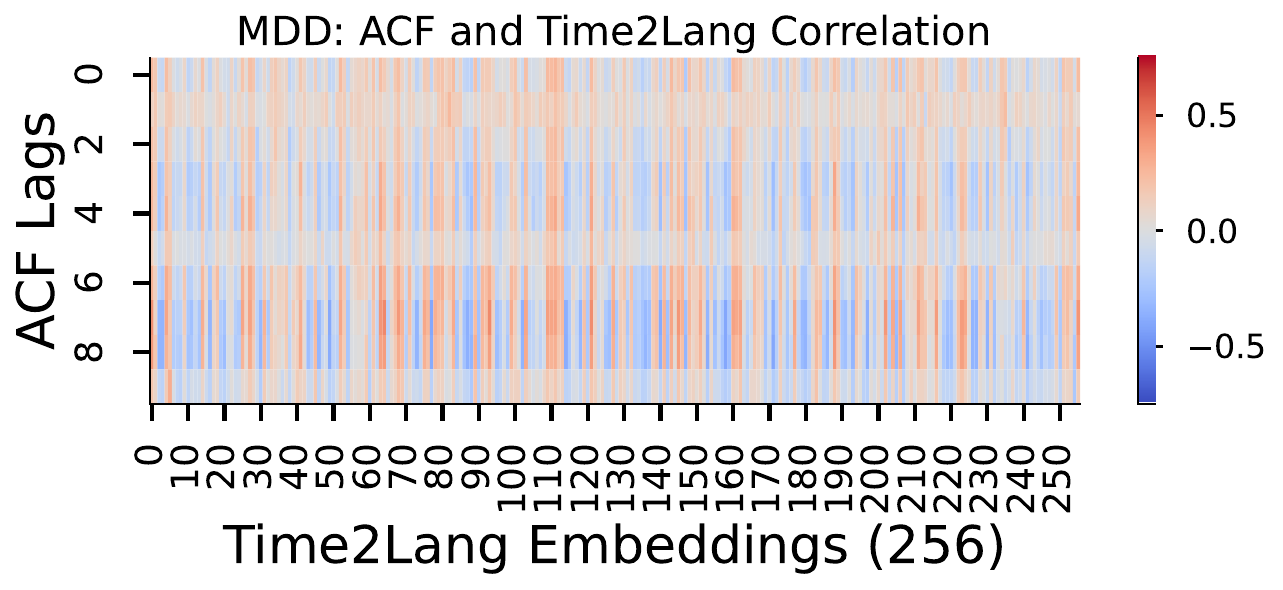}}
    \subfigure[]{\includegraphics[width=0.15\linewidth]{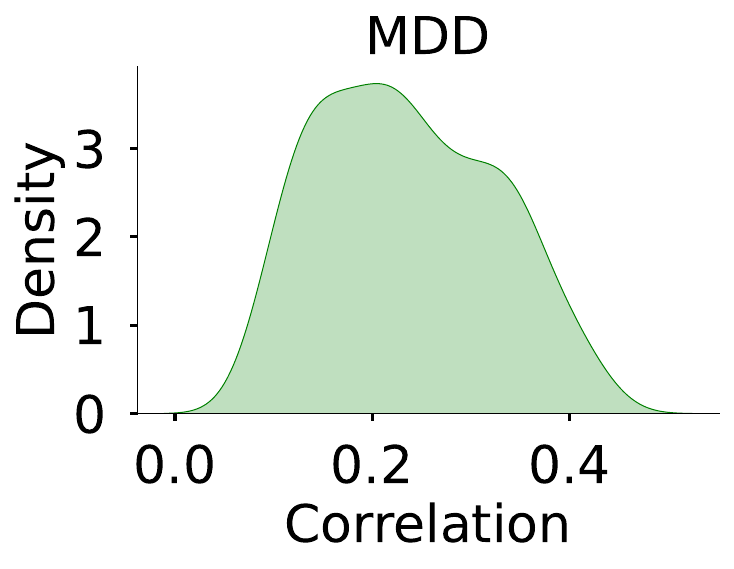}}
    \caption{\textbf{Comparing embeddings' temporal dynamics.} The Spearman rank correlation between Time2Lang embeddings and the Auto-Correlation Factor (ACF) of the original sensing data. Subfigures (a) and (c) illustrate the correlation between the ACF of conversation and step count data with their respective embeddings, while subfigures (b) and (d) display the distribution of the strongest correlation values across all 256 embedding dimensions.}
    \label{fig:correlation}
\end{figure*}

We evaluate the impact of synthetic data size and number of periodicity classes on downstream performance. Specifically, we consider dataset sizes of 50K, 100K, and 200K samples and periodicity class counts of 3 (30 to 90), 6 (30 to 180), and 9 (30 to 270), resulting in nine combinations. As shown in Figure \ref{fig:effect_of_pre_training}, pre-training with 200K samples and 6 periodicity classes achieved the best performance across all cases. Among dataset sizes, performance improves as data increases from 50K to 100K, but the difference between 100K and 200K is marginal. Similarly, increasing periodicity classes from 3 to 6 generally enhances performance. However, further increasing the classes to 9 does not yield additional improvements, regardless of dataset size. We attribute this plateau to the model’s parameter capacity, as a 9-class classification problem demands a more complex network, exceeding the capacity of the trainable parameters.

\subsection{Ablation Study: Architecture} \label{sec:architecure}
Our ablation studies investigates the influence of Time2Lang's architectural elements. Specifically, we analyze the role of residual connections, variations in $f$ and $g$, and the impact of different LLMs (Appendix~\S\ref{apd:ablation}). The inclusion of a residual connection ($\mathbf{z^c} \rightarrow \mathbf{z^m}$) generally yielded superior results, highlighting its importance for preserving beneficial information transfer (Table \ref{tab:residual}; Appendix~\S\ref{apd:residual_ablation}). Furthermore, we explored the effects of wider and deeper ResNet input encoders ($f$) and larger projection embedding sizes ($g$)(Appendix~\S\ref{apd:architecture_ablation}). In contrast to the base Time2Lang configuration (32 initial filters, 256 embedding size), the wider Time2Lang-L (64 base filters, 512 embedding size) demonstrated improved performance. Notably, in predicting flourishing from conversation data, Time2Lang-L achieved an AUROC of 0.80 and an AUPRC of 0.85, representing gains of +0.09 and +0.11 over Time2Lang (Table \ref{tab:architecture_ablations}). Interestingly, increasing the depth of the ResNet encoder did not lead to performance enhancements, suggesting that wider architectures may be more advantageous. Finally, our evaluation of alternative LLMs within the Time2Lang framework (Table \ref{tab:llm_backbones}; Appendix~\S\ref{apd:llm_ablation}) indicated that LLaMA 3.2 performed the best in general. However, further work is needed to integrate Gemma 3 and OLMo more effectively.

\subsection{Case Study: Correlation of Embeddings and ACF} 
\label{sec:case_study}
We aimed to evaluate whether, after reprogramming through Time2Lang, the LLM outputs effectively capture fundamental time series characteristics. Thus, we computed the Spearman rank correlation between the Auto-Correlation Factor (ACF) of the original time series and the embeddings. First, we calculated the ACF with 10 lags from the original sensing data. Then, we computed the pairwise correlation between each ACF lag and the corresponding Time2Lang embedding across the samples. The heatmaps in Figure \ref{fig:correlation} (a) and (c) show that Time2Lang embeddings capture both positive and negative correlations across conversation and step count data. Notably, we observe stronger correlations at higher time lags (7–9), suggesting that Time2Lang embeddings retain information extending beyond recent past observations. To further analyze the strength of these correlations, we identified the highest correlation value for each embedding dimension at any lag. The distribution of these values, shown in Figure \ref{fig:correlation} (b) and (d), indicates that for conversation data, 174 dimensions ($\sim$68\%) have at least one lag where the correlation exceeds 0.3, signifying a moderate correlation. Similarly, for step count data, 70 dimensions ($\sim$27\%) demonstrate a moderate correlation greater than 0.3.

\section{Discussion}

\subsection{Findings} \label{sec:findings}

Time2Lang achieved an AUROC=0.57 and AUPRC=0.73 for predicting daily depression based on step count and an AUROC=0.71 and AUPRC=0.74 for predicting flourishing. In the context of mental health, its depression prediction performance aligns with findings from previous studies \citep{xu2023globem}. Among prompting approaches, incorporating specific domain knowledge enhances performance, consistent with prior observations by \citet{kim2024health}. However, Time2Lang consistently outperforms prompting across both metrics and tasks. The LLaMA + Chronos approach showed markedly lower performance than Chronos alone, indicating that information is lost when integrating two distinct models. Our goal is to minimize performance loss during translation when leveraging Chronos. In this regard, Time2Lang outperforms LLaMA + Chronos, demonstrating that learning a direct mapping yields better results than simply combining embeddings. Ultimately, Time2Lang achieves both marginal and notable improvements over Chronos, showcasing that a TFM can be integrated with an LLM by minimizing information loss across these two distinct domains.

Through our efficiency analysis, we find that Time2Lang's inference time is steady when increasing input length. However, we notice that LLaMA Prompting's inference time increases with longer duration sequences primarily because of more tokens. The faster inference time of Time2Lang can be attributed to the TFM that produces fixed sized embeddings, irrespective of input length. Our correlation analysis reveals that after mapping, the LLM outputs from Time2Lang exhibit moderate to high correlations with ACF, indicating that our method effectively captures key time series properties. We attribute this capability to our self-supervised pre-training, which is designed to predict periodicity. Interestingly, this pre-training on synthetic data, generalizes to real-world datasets, demonstrating the robustness of our approach.

\subsection{Implications}

Our work has significant implications for the development of personalized mental health LLMs and context-aware interventions. It represents an essential first step toward integrating sensing data into LLM-driven mental health systems, enabling them to provide critical, user-specific insights. Prior research has demonstrated that mental health symptoms fluctuate significantly within a single day and across different populations \citep{lim2018prevalence, crowe2019intra}. A personalized system that continuously monitors behavioral sensing data can identify individual-specific patterns, offering a more nuanced understanding of mental health variations. Unlike traditional check-ups, which provide only periodic assessments, continuous monitoring enables dynamic, real-time recommendations tailored to the user’s needs. However, to deliver accurate and effective mental health recommendations, the first step is ensuring that the model can comprehend and process sensing data efficiently. Scalability and computational efficiency are crucial for handling continuous sensing streams. As discussed in Sections \ref{sec:results} and \ref{sec:findings}, the Time2Lang approach minimizes information loss while achieving performance comparable to a TFM, improving efficiency, and establishing strong connections to fundamental time series characteristics. These results show that fusing two FMs from distinct domains can lead to meaningful representations, consequently paving the way for future work that may enable real-time generation and dynamic mental health interventions for users.


\subsection{Limitations \& Future Work}

We acknowledge the limitations of our work, particularly its reliance on uni-modal time series data, as Time2Lang (Chronos) processes a single modality. Extending it to multi-modal data is a promising direction that would enhance Time2Lang's ability to handle richer data. While we demonstrate Time2Lang’s effectiveness for mental health, its potential extends to other areas including physiological sensing, industrial sensors, and environmental monitoring. Therefore, we plan to evaluate our method on more diverse datasets to evaluate generalizability in the future.

The selection of Chronos and LLaMA is motivated by their fixed-length embeddings, relation to language modeling, and practical utility, as discussed in Section \ref{sec:Time2Lang}. However, different tasks or modalities may benefit from alternative pre-trained models. Rather than prescribing a one-size-fits-all approach, we aim to provide a new perspective on leveraging LLMs for sensing data analysis. Future work should systematically evaluate modality interactions and model suitability.

Designing computationally efficient reasoning methods for behavioral sensing in mental health remains challenging. Prior research suggests reasoning is strongly tied to scaling \citep{moon2024anymal, girdhar2023imagebind, spathis2024first}. While our approach emphasizes efficiency, its reasoning capabilities remain limited. However, Time2Lang provides a foundation for integrating time series data into LLMs, which we aim to extend for more complex reasoning and generative tasks. Ultimately, our goal is to develop a personalized LLM that understands individual user behavior through daily data, enabling more tailored and insightful interactions.

\vspace{-0.2cm}
\section{Conclusion}
In this work, we propose Time2Lang, a framework that integrates sensing data with LLMs by learning a direct mapping from a TFM (Chronos) to an LLM (LLaMA).  This approach overcomes the limitations of text-based prompting. Through extensive evaluation on real-world longitudinal datasets for tasks such as daily depression prediction and flourishing classification, we demonstrate that Time2Lang positively integrates a TFM and an LLM while maintaining stable inference times across varying input lengths. Furthermore, our correlation analysis highlights Time2Lang effectively reprograms LLM outputs to capture fundamental time-series properties such as auto-correlation. Ultimately, Time2Lang enhances efficiency, accuracy, and time-series comprehension compared to traditional prompting methods. By bridging the gap between behavioral sensing and LLM-based inference, Time2Lang represents a significant step toward enhancing AI-driven mental health assessment. Future work will explore its extension to reasoning with multi-modal sensing data, further improving its applicability in personalized mental health monitoring and intervention systems.

\begin{acks}
This work was supported by the National Institute of Mental Health (NIMH) under award number R01MH123482-01 and the Evergreen project at Dartmouth. It is important to note that the opinions expressed herein are those of the authors and do not necessarily reflect the official policy or position of the NIMH. We also state that the NIMH had no influence on the study's design, data collection, analysis, interpretation of the data, or the writing of this report.
\end{acks}

\bibliography{chil-sample}

\begin{thebibliography}{50}
\providecommand{\natexlab}[1]{#1}
\providecommand{\url}[1]{\texttt{#1}}
\expandafter\ifx\csname urlstyle\endcsname\relax
  \providecommand{\doi}[1]{doi: #1}\else
  \providecommand{\doi}{doi: \begingroup \urlstyle{rm}\Url}\fi

\bibitem[Ansari et~al.(2024)Ansari, Stella, Turkmen, Zhang, Mercado, Shen, Shchur, Rangapuram, Arango, Kapoor, et~al.]{ansari2024chronos}
Abdul~Fatir Ansari, Lorenzo Stella, Caner Turkmen, Xiyuan Zhang, Pedro Mercado, Huibin Shen, Oleksandr Shchur, Syama~Sundar Rangapuram, Sebastian~Pineda Arango, Shubham Kapoor, et~al.
\newblock Chronos: Learning the language of time series.
\newblock \emph{arXiv preprint arXiv:2403.07815}, 2024.

\bibitem[Belyaeva et~al.(2023)Belyaeva, Cosentino, Hormozdiari, Eswaran, Shetty, Corrado, Carroll, McLean, and Furlotte]{belyaeva2023multimodal}
Anastasiya Belyaeva, Justin Cosentino, Farhad Hormozdiari, Krish Eswaran, Shravya Shetty, Greg Corrado, Andrew Carroll, Cory~Y McLean, and Nicholas~A Furlotte.
\newblock Multimodal llms for health grounded in individual-specific data.
\newblock In \emph{Workshop on Machine Learning for Multimodal Healthcare Data}, pages 86--102. Springer, 2023.

\bibitem[Bhattacharya et~al.(2024)Bhattacharya, Majethia, Choube, and Mishra]{bhattacharya2024imputation}
Sohini Bhattacharya, Rahul Majethia, Akshat Choube, and Varun Mishra.
\newblock Imputation strategies for longitudinal behavioral studies: Predicting depression using globem datasets.
\newblock In \emph{Companion of the 2024 on ACM International Joint Conference on Pervasive and Ubiquitous Computing}, pages 736--742, 2024.

\bibitem[Bizzozero-Peroni et~al.(2024)Bizzozero-Peroni, D{\'\i}az-Go{\~n}i, Jim{\'e}nez-L{\'o}pez, Rodr{\'\i}guez-Guti{\'e}rrez, Sequ{\'\i}-Dom{\'\i}nguez, de~Arenas-Arroyo, L{\'o}pez-Gil, Mart{\'\i}nez-Vizca{\'\i}no, and Mesas]{bizzozero2024daily}
Bruno Bizzozero-Peroni, Valentina D{\'\i}az-Go{\~n}i, Estela Jim{\'e}nez-L{\'o}pez, Eva Rodr{\'\i}guez-Guti{\'e}rrez, Irene Sequ{\'\i}-Dom{\'\i}nguez, Sergio~N{\'u}{\~n}ez de~Arenas-Arroyo, Jos{\'e}~Francisco L{\'o}pez-Gil, Vicente Mart{\'\i}nez-Vizca{\'\i}no, and Arthur~Eumann Mesas.
\newblock Daily step count and depression in adults: A systematic review and meta-analysis.
\newblock \emph{JAMA Network Open}, 7\penalty0 (12):\penalty0 e2451208--e2451208, 2024.

\bibitem[Bommasani et~al.(2021)Bommasani, Hudson, Adeli, Altman, Arora, von Arx, Bernstein, Bohg, Bosselut, Brunskill, et~al.]{bommasani2021opportunities}
Rishi Bommasani, Drew~A Hudson, Ehsan Adeli, Russ Altman, Simran Arora, Sydney von Arx, Michael~S Bernstein, Jeannette Bohg, Antoine Bosselut, Emma Brunskill, et~al.
\newblock On the opportunities and risks of foundation models.
\newblock \emph{arXiv preprint arXiv:2108.07258}, 2021.

\bibitem[Cai et~al.(2024)Cai, Srinivasan, Goswami, Choudhry, and Dubrawski]{cai2024jolt}
Yifu Cai, Arvind Srinivasan, Mononito Goswami, Arjun Choudhry, and Artur Dubrawski.
\newblock Jolt: jointly learned representations of language and time-series for clinical time-series interpretation (student abstract).
\newblock In \emph{Proceedings of the AAAI Conference on Artificial Intelligence}, volume~38, pages 23447--23448, 2024.

\bibitem[Chen(2024)]{chen2024model}
Pin-Yu Chen.
\newblock Model reprogramming: Resource-efficient cross-domain machine learning.
\newblock In \emph{Proceedings of the AAAI Conference on Artificial Intelligence}, volume~38, pages 22584--22591, 2024.

\bibitem[Conmy et~al.(2023)Conmy, Mavor-Parker, Lynch, Heimersheim, and Garriga-Alonso]{conmy2023towards}
Arthur Conmy, Augustine Mavor-Parker, Aengus Lynch, Stefan Heimersheim, and Adri{\`a} Garriga-Alonso.
\newblock Towards automated circuit discovery for mechanistic interpretability.
\newblock \emph{Advances in Neural Information Processing Systems}, 36:\penalty0 16318--16352, 2023.

\bibitem[Cosentino et~al.(2024)Cosentino, Belyaeva, Liu, Furlotte, Yang, Lee, Schenck, Patel, Cui, Schneider, et~al.]{cosentino2024towards}
Justin Cosentino, Anastasiya Belyaeva, Xin Liu, Nicholas~A Furlotte, Zhun Yang, Chace Lee, Erik Schenck, Yojan Patel, Jian Cui, Logan~Douglas Schneider, et~al.
\newblock Towards a personal health large language model.
\newblock \emph{arXiv preprint arXiv:2406.06474}, 2024.

\bibitem[Crowe et~al.(2019)Crowe, Daly, Delaney, Carroll, and Malone]{crowe2019intra}
Eimear Crowe, Michael Daly, Liam Delaney, Susan Carroll, and Kevin~M Malone.
\newblock The intra-day dynamics of affect, self-esteem, tiredness, and suicidality in major depression.
\newblock \emph{Psychiatry Research}, 279:\penalty0 98--108, 2019.

\bibitem[Deldari et~al.(2024)Deldari, Spathis, Malekzadeh, Kawsar, Salim, and Mathur]{deldari2024crossl}
Shohreh Deldari, Dimitris Spathis, Mohammad Malekzadeh, Fahim Kawsar, Flora~D Salim, and Akhil Mathur.
\newblock Crossl: Cross-modal self-supervised learning for time-series through latent masking.
\newblock In \emph{Proceedings of the 17th ACM International Conference on Web Search and Data Mining}, pages 152--160, 2024.

\bibitem[Diener et~al.(2010)Diener, Wirtz, Tov, Kim-Prieto, Choi, Oishi, and Biswas-Diener]{diener2010new}
Ed~Diener, Derrick Wirtz, William Tov, Chu Kim-Prieto, Dong-won Choi, Shigehiro Oishi, and Robert Biswas-Diener.
\newblock New well-being measures: Short scales to assess flourishing and positive and negative feelings.
\newblock \emph{Social indicators research}, 97:\penalty0 143--156, 2010.

\bibitem[Dong et~al.(2022)Dong, Li, Dai, Zheng, Ma, Li, Xia, Xu, Wu, Liu, et~al.]{dong2022survey}
Qingxiu Dong, Lei Li, Damai Dai, Ce~Zheng, Jingyuan Ma, Rui Li, Heming Xia, Jingjing Xu, Zhiyong Wu, Tianyu Liu, et~al.
\newblock A survey on in-context learning.
\newblock \emph{arXiv preprint arXiv:2301.00234}, 2022.

\bibitem[Dubey et~al.(2024)Dubey, Jauhri, Pandey, Kadian, Al-Dahle, Letman, Mathur, Schelten, Yang, Fan, et~al.]{dubey2024llama}
Abhimanyu Dubey, Abhinav Jauhri, Abhinav Pandey, Abhishek Kadian, Ahmad Al-Dahle, Aiesha Letman, Akhil Mathur, Alan Schelten, Amy Yang, Angela Fan, et~al.
\newblock The llama 3 herd of models.
\newblock \emph{arXiv preprint arXiv:2407.21783}, 2024.

\bibitem[Ghosh et~al.(2024)Ghosh, Kumar, Seth, Evuru, Tyagi, Sakshi, Nieto, Duraiswami, and Manocha]{ghosh2024gama}
Sreyan Ghosh, Sonal Kumar, Ashish Seth, Chandra Kiran~Reddy Evuru, Utkarsh Tyagi, S~Sakshi, Oriol Nieto, Ramani Duraiswami, and Dinesh Manocha.
\newblock Gama: A large audio-language model with advanced audio understanding and complex reasoning abilities.
\newblock \emph{arXiv preprint arXiv:2406.11768}, 2024.

\bibitem[Girdhar et~al.(2023)Girdhar, El-Nouby, Liu, Singh, Alwala, Joulin, and Misra]{girdhar2023imagebind}
Rohit Girdhar, Alaaeldin El-Nouby, Zhuang Liu, Mannat Singh, Kalyan~Vasudev Alwala, Armand Joulin, and Ishan Misra.
\newblock Imagebind: One embedding space to bind them all.
\newblock In \emph{Proceedings of the IEEE/CVF Conference on Computer Vision and Pattern Recognition}, pages 15180--15190, 2023.

\bibitem[Goswami et~al.(2024)Goswami, Szafer, Choudhry, Cai, Li, and Dubrawski]{goswami2024moment}
Mononito Goswami, Konrad Szafer, Arjun Choudhry, Yifu Cai, Shuo Li, and Artur Dubrawski.
\newblock Moment: A family of open time-series foundation models.
\newblock \emph{arXiv preprint arXiv:2402.03885}, 2024.

\bibitem[Groeneveld et~al.(2024)Groeneveld, Beltagy, Walsh, Bhagia, Kinney, Tafjord, Jha, Ivison, Magnusson, Wang, et~al.]{groeneveld2024olmo}
Dirk Groeneveld, Iz~Beltagy, Pete Walsh, Akshita Bhagia, Rodney Kinney, Oyvind Tafjord, Ananya~Harsh Jha, Hamish Ivison, Ian Magnusson, Yizhong Wang, et~al.
\newblock Olmo: Accelerating the science of language models.
\newblock \emph{arXiv preprint arXiv:2402.00838}, 2024.

\bibitem[Gruver et~al.(2024)Gruver, Finzi, Qiu, and Wilson]{gruver2024large}
Nate Gruver, Marc Finzi, Shikai Qiu, and Andrew~G Wilson.
\newblock Large language models are zero-shot time series forecasters.
\newblock \emph{Advances in Neural Information Processing Systems}, 36, 2024.

\bibitem[Gu et~al.(2019)Gu, Pandit, Saraee, Nordahl, Ellis, and Betke]{gu2019home}
Yiwen Gu, Shreya Pandit, Elham Saraee, Timothy Nordahl, Terry Ellis, and Margrit Betke.
\newblock Home-based physical therapy with an interactive computer vision system.
\newblock In \emph{Proceedings of the IEEE/CVF International Conference on Computer Vision Workshops}, pages 0--0, 2019.

\bibitem[Han et~al.(2023)Han, Adams, Papaioannou, Grundmann, Oberhauser, L{\"o}ser, Truhn, and Bressem]{han2023medalpaca}
Tianyu Han, Lisa~C Adams, Jens-Michalis Papaioannou, Paul Grundmann, Tom Oberhauser, Alexander L{\"o}ser, Daniel Truhn, and Keno~K Bressem.
\newblock Medalpaca--an open-source collection of medical conversational ai models and training data.
\newblock \emph{arXiv preprint arXiv:2304.08247}, 2023.

\bibitem[He et~al.(2023)He, Mao, Lin, Ruan, Lan, Feng, and Cambria]{he2023survey}
Kai He, Rui Mao, Qika Lin, Yucheng Ruan, Xiang Lan, Mengling Feng, and Erik Cambria.
\newblock A survey of large language models for healthcare: from data, technology, and applications to accountability and ethics.
\newblock \emph{arXiv preprint arXiv:2310.05694}, 2023.

\bibitem[Kaplan et~al.(2020)Kaplan, McCandlish, Henighan, Brown, Chess, Child, Gray, Radford, Wu, and Amodei]{kaplan2020scaling}
Jared Kaplan, Sam McCandlish, Tom Henighan, Tom~B Brown, Benjamin Chess, Rewon Child, Scott Gray, Alec Radford, Jeffrey Wu, and Dario Amodei.
\newblock Scaling laws for neural language models.
\newblock \emph{arXiv preprint arXiv:2001.08361}, 2020.

\bibitem[Kim et~al.(2024)Kim, Xu, McDuff, Breazeal, and Park]{kim2024health}
Yubin Kim, Xuhai Xu, Daniel McDuff, Cynthia Breazeal, and Hae~Won Park.
\newblock Health-llm: Large language models for health prediction via wearable sensor data.
\newblock \emph{arXiv preprint arXiv:2401.06866}, 2024.

\bibitem[Kroenke et~al.(2001)Kroenke, Spitzer, and Williams]{kroenke2001phq}
Kurt Kroenke, Robert~L Spitzer, and Janet~BW Williams.
\newblock The phq-9: validity of a brief depression severity measure.
\newblock \emph{Journal of general internal medicine}, 16\penalty0 (9):\penalty0 606--613, 2001.

\bibitem[Lim et~al.(2018)Lim, Tam, Lu, Ho, Zhang, and Ho]{lim2018prevalence}
Grace~Y Lim, Wilson~W Tam, Yanxia Lu, Cyrus~S Ho, Melvyn~W Zhang, and Roger~C Ho.
\newblock Prevalence of depression in the community from 30 countries between 1994 and 2014.
\newblock \emph{Scientific reports}, 8\penalty0 (1):\penalty0 2861, 2018.

\bibitem[Liu et~al.(2023{\natexlab{a}})Liu, Yuan, Fu, Jiang, Hayashi, and Neubig]{liu2023pre}
Pengfei Liu, Weizhe Yuan, Jinlan Fu, Zhengbao Jiang, Hiroaki Hayashi, and Graham Neubig.
\newblock Pre-train, prompt, and predict: A systematic survey of prompting methods in natural language processing.
\newblock \emph{ACM Computing Surveys}, 55\penalty0 (9):\penalty0 1--35, 2023{\natexlab{a}}.

\bibitem[Liu et~al.(2025)Liu, Liu, Yang, Jiang, Cui, Zhang, Wang, Tao, Sun, Song, et~al.]{liu2025generalist}
Xiaohong Liu, Hao Liu, Guoxing Yang, Zeyu Jiang, Shuguang Cui, Zhaoze Zhang, Huan Wang, Liyuan Tao, Yongchang Sun, Zhu Song, et~al.
\newblock A generalist medical language model for disease diagnosis assistance.
\newblock \emph{Nature Medicine}, pages 1--11, 2025.

\bibitem[Liu et~al.(2023{\natexlab{b}})Liu, McDuff, Kovacs, Galatzer-Levy, Sunshine, Zhan, Poh, Liao, Di~Achille, and Patel]{liu2023large}
Xin Liu, Daniel McDuff, Geza Kovacs, Isaac Galatzer-Levy, Jacob Sunshine, Jiening Zhan, Ming-Zher Poh, Shun Liao, Paolo Di~Achille, and Shwetak Patel.
\newblock Large language models are few-shot health learners.
\newblock \emph{arXiv preprint arXiv:2305.15525}, 2023{\natexlab{b}}.

\bibitem[Moon et~al.(2024)Moon, Madotto, Lin, Nagarajan, Smith, Jain, Yeh, Murugesan, Heidari, Liu, et~al.]{moon2024anymal}
Seungwhan Moon, Andrea Madotto, Zhaojiang Lin, Tushar Nagarajan, Matt Smith, Shashank Jain, Chun-Fu Yeh, Prakash Murugesan, Peyman Heidari, Yue Liu, et~al.
\newblock Anymal: An efficient and scalable any-modality augmented language model.
\newblock In \emph{Proceedings of the 2024 Conference on Empirical Methods in Natural Language Processing: Industry Track}, pages 1314--1332, 2024.

\bibitem[Neekhara et~al.(2022)Neekhara, Hussain, Du, Dubnov, Koushanfar, and McAuley]{neekhara2022cross}
Paarth Neekhara, Shehzeen Hussain, Jinglong Du, Shlomo Dubnov, Farinaz Koushanfar, and Julian McAuley.
\newblock Cross-modal adversarial reprogramming.
\newblock In \emph{Proceedings of the IEEE/CVF Winter Conference on Applications of Computer Vision}, pages 2427--2435, 2022.

\bibitem[Nepal et~al.(2024)Nepal, Liu, Pillai, Wang, Vojdanovski, Huckins, Rogers, Meyer, and Campbell]{nepal2024capturing}
Subigya Nepal, Wenjun Liu, Arvind Pillai, Weichen Wang, Vlado Vojdanovski, Jeremy~F Huckins, Courtney Rogers, Meghan~L Meyer, and Andrew~T Campbell.
\newblock Capturing the college experience: A four-year mobile sensing study of mental health, resilience and behavior of college students during the pandemic.
\newblock \emph{Proceedings of the ACM on Interactive, Mobile, Wearable and Ubiquitous Technologies}, 8\penalty0 (1):\penalty0 1--37, 2024.

\bibitem[Pillai et~al.(2024)Pillai, Spathis, Kawsar, and Malekzadeh]{pillai2024papagei}
Arvind Pillai, Dimitris Spathis, Fahim Kawsar, and Mohammad Malekzadeh.
\newblock Papagei: Open foundation models for optical physiological signals.
\newblock \emph{arXiv preprint arXiv:2410.20542}, 2024.

\bibitem[Price et~al.(2023)Price, Langener, Heinz, Mackin, Nemesure, Collins, Griffin, Pillai, Nepal, Lekkas, et~al.]{price2023predicting}
George Price, Anna Langener, Michael~V Heinz, Daniel Mackin, Matthew~D Nemesure, Amanda~C Collins, Tess Griffin, Arvind Pillai, Subigya Nepal, Damien Lekkas, et~al.
\newblock Predicting weekly variability in depressive symptoms among individuals diagnosed with major depressive disorder using deep learning and passively gathered movement data.
\newblock 2023.

\bibitem[Radford et~al.(2021)Radford, Kim, Hallacy, Ramesh, Goh, Agarwal, Sastry, Askell, Mishkin, Clark, et~al.]{radford2021learning}
Alec Radford, Jong~Wook Kim, Chris Hallacy, Aditya Ramesh, Gabriel Goh, Sandhini Agarwal, Girish Sastry, Amanda Askell, Pamela Mishkin, Jack Clark, et~al.
\newblock Learning transferable visual models from natural language supervision.
\newblock In \emph{International conference on machine learning}, pages 8748--8763. PMLR, 2021.

\bibitem[Schulz et~al.(2018)Schulz, Speekenbrink, and Krause]{schulz2018tutorial}
Eric Schulz, Maarten Speekenbrink, and Andreas Krause.
\newblock A tutorial on gaussian process regression: Modelling, exploring, and exploiting functions.
\newblock \emph{Journal of mathematical psychology}, 85:\penalty0 1--16, 2018.

\bibitem[Singhal et~al.(2025)Singhal, Tu, Gottweis, Sayres, Wulczyn, Amin, Hou, Clark, Pfohl, Cole-Lewis, et~al.]{singhal2025toward}
Karan Singhal, Tao Tu, Juraj Gottweis, Rory Sayres, Ellery Wulczyn, Mohamed Amin, Le~Hou, Kevin Clark, Stephen~R Pfohl, Heather Cole-Lewis, et~al.
\newblock Toward expert-level medical question answering with large language models.
\newblock \emph{Nature Medicine}, pages 1--8, 2025.

\bibitem[Spathis and Kawsar(2024)]{spathis2024first}
Dimitris Spathis and Fahim Kawsar.
\newblock The first step is the hardest: Pitfalls of representing and tokenizing temporal data for large language models.
\newblock \emph{Journal of the American Medical Informatics Association}, 31\penalty0 (9):\penalty0 2151--2158, 2024.

\bibitem[Team et~al.(2025)Team, Kamath, Ferret, Pathak, Vieillard, Merhej, Perrin, Matejovicova, Ram{\'e}, Rivi{\`e}re, et~al.]{team2025gemma}
Gemma Team, Aishwarya Kamath, Johan Ferret, Shreya Pathak, Nino Vieillard, Ramona Merhej, Sarah Perrin, Tatiana Matejovicova, Alexandre Ram{\'e}, Morgane Rivi{\`e}re, et~al.
\newblock Gemma 3 technical report.
\newblock \emph{arXiv preprint arXiv:2503.19786}, 2025.

\bibitem[Vinod et~al.(2023)Vinod, Chen, and Das]{vinod2023reprogramming}
Ria Vinod, Pin-Yu Chen, and Payel Das.
\newblock Reprogramming pretrained language models for protein sequence representation learning.
\newblock \emph{arXiv preprint arXiv:2301.02120}, 2023.

\bibitem[Wang et~al.(2014)Wang, Chen, Chen, Li, Harari, Tignor, Zhou, Ben-Zeev, and Campbell]{wang2014studentlife}
Rui Wang, Fanglin Chen, Zhenyu Chen, Tianxing Li, Gabriella Harari, Stefanie Tignor, Xia Zhou, Dror Ben-Zeev, and Andrew~T Campbell.
\newblock Studentlife: assessing mental health, academic performance and behavioral trends of college students using smartphones.
\newblock In \emph{Proceedings of the 2014 ACM international joint conference on pervasive and ubiquitous computing}, pages 3--14, 2014.

\bibitem[Xie et~al.(2024)Xie, Chen, Chen, Peng, Hu, Lin, Peng, Huang, Zhang, Keloth, et~al.]{xie2024me}
Qianqian Xie, Qingyu Chen, Aokun Chen, Cheng Peng, Yan Hu, Fongci Lin, Xueqing Peng, Jimin Huang, Jeffrey Zhang, Vipina Keloth, et~al.
\newblock Me-llama: Medical foundation large language models for comprehensive text analysis and beyond.
\newblock 2024.

\bibitem[Xu et~al.(2024{\natexlab{a}})Xu, Han, Yang, Li, and Srivastava]{xu2024penetrative}
Huatao Xu, Liying Han, Qirui Yang, Mo~Li, and Mani Srivastava.
\newblock Penetrative ai: Making llms comprehend the physical world.
\newblock In \emph{Proceedings of the 25th International Workshop on Mobile Computing Systems and Applications}, pages 1--7, 2024{\natexlab{a}}.

\bibitem[Xu et~al.(2023)Xu, Liu, Zhang, Wang, Nepal, Sefidgar, Seo, Kuehn, Huckins, Morris, et~al.]{xu2023globem}
Xuhai Xu, Xin Liu, Han Zhang, Weichen Wang, Subigya Nepal, Yasaman Sefidgar, Woosuk Seo, Kevin~S Kuehn, Jeremy~F Huckins, Margaret~E Morris, et~al.
\newblock Globem: cross-dataset generalization of longitudinal human behavior modeling.
\newblock \emph{Proceedings of the ACM on Interactive, Mobile, Wearable and Ubiquitous Technologies}, 6\penalty0 (4):\penalty0 1--34, 2023.

\bibitem[Xu et~al.(2024{\natexlab{b}})Xu, Yao, Dong, Gabriel, Yu, Hendler, Ghassemi, Dey, and Wang]{xu2024mental}
Xuhai Xu, Bingsheng Yao, Yuanzhe Dong, Saadia Gabriel, Hong Yu, James Hendler, Marzyeh Ghassemi, Anind~K Dey, and Dakuo Wang.
\newblock Mental-llm: Leveraging large language models for mental health prediction via online text data.
\newblock \emph{Proceedings of the ACM on Interactive, Mobile, Wearable and Ubiquitous Technologies}, 8\penalty0 (1):\penalty0 1--32, 2024{\natexlab{b}}.

\bibitem[Xue and Salim(2023)]{xue2023promptcast}
Hao Xue and Flora~D Salim.
\newblock Promptcast: A new prompt-based learning paradigm for time series forecasting.
\newblock \emph{IEEE Transactions on Knowledge and Data Engineering}, 2023.

\bibitem[Yang et~al.(2021)Yang, Tsai, and Chen]{yang2021voice2series}
Chao-Han~Huck Yang, Yun-Yun Tsai, and Pin-Yu Chen.
\newblock Voice2series: Reprogramming acoustic models for time series classification.
\newblock In \emph{International conference on machine learning}, pages 11808--11819. PMLR, 2021.

\bibitem[Yang et~al.(2022)Yang, Liu, Wu, Borac, Katabi, Poh, and McDuff]{yang2022simper}
Yuzhe Yang, Xin Liu, Jiang Wu, Silviu Borac, Dina Katabi, Ming-Zher Poh, and Daniel McDuff.
\newblock Simper: Simple self-supervised learning of periodic targets.
\newblock \emph{arXiv preprint arXiv:2210.03115}, 2022.

\bibitem[Yoon et~al.(2024)Yoon, Tolera, Gong, Lee, and Lee]{yoon2024my}
Hyungjun Yoon, Biniyam~Aschalew Tolera, Taesik Gong, Kimin Lee, and Sung-Ju Lee.
\newblock By my eyes: Grounding multimodal large language models with sensor data via visual prompting.
\newblock \emph{arXiv preprint arXiv:2407.10385}, 2024.

\bibitem[Zamfirescu-Pereira et~al.(2023)Zamfirescu-Pereira, Wong, Hartmann, and Yang]{zamfirescu2023johnny}
JD~Zamfirescu-Pereira, Richmond~Y Wong, Bjoern Hartmann, and Qian Yang.
\newblock Why johnny can’t prompt: how non-ai experts try (and fail) to design llm prompts.
\newblock In \emph{Proceedings of the 2023 CHI Conference on Human Factors in Computing Systems}, pages 1--21, 2023.

\end{thebibliography}
\newpage

\appendix

\section{Ethical Considerations} \label{apd:ethics}

Our study adheres to ethical guidelines for participant safety, privacy, and regulatory compliance, as approved by the Institutional Review Board. To prioritize participant well-being, three clinical psychologists/psychiatrists were available to address questions and provide feedback. Additionally, our application automatically alerts the study team in cases of severe depression or active suicidality. Our research on Time2Lang follows data privacy regulations. For example, the data is stored in a 2FA server with access only to approved researchers even among the core team. We acknowledge potential biases in the training data, as most participants are white, female, and based in the United States—reflecting the prevalence of MDD in this demographic. Although Time2Lang offers significant potential for non-invasive health monitoring, we acknowledge the possibility of misuse, including unauthorized surveillance of health data, biased decision-making in insurance and employment, inequitable credit assessments, and the exploitation of personal health information for targeted advertising. We strongly emphasize the importance of using this technology responsibly within healthcare settings. Our analysis follows established research ethics guidelines. We actively support cross-disciplinary collaboration to identify and minimize potential risks, fostering the ethical advancement of AI in healthcare.

\section{Architecture \& Hyperparameters} \label{apd:architecture}
The input encoder ($f$) comprises 13 convolutional layers (Table \ref{tab:resnet}): an initial layer, followed by six 2-layer blocks (Tables \ref{tab:resnet_block1} and \ref{tab:resnet_block2}). The input channels are 768, and we begin with 32 base filters, using a kernel size of 3 and stride of 2. The projection layer consists of two fully connected layers with 768 and 256 features, respectively. Each layer is followed by batch normalization and ReLU activation to ensure stable training and improved representation learning.

\begin{table}[h]
    \centering
    \caption{ResNet-Style CNN}
    \begin{tabular}{l}
    \toprule
        \textbf{Layer}\\
        \midrule
        Conv1D \\
        BatchNorm \\
        ReLU \\
        (Basic Block Type 1) x 1 \\
        (Basic Block Type 2) x 5 \\
        BatchNorm \\
        ReLU\\
        \bottomrule
    \end{tabular}
    \label{tab:resnet}
\end{table}

\begin{table}[h]
\centering
\begin{minipage}{0.4\linewidth}
    \centering
    \caption{Basic Block Type 1}
    \begin{tabular}{@{}l@{}}
    \toprule
    \textbf{Layer} \\ \midrule
    Conv1D \\
    BatchNorm \\
    ReLU \\
    Dropout \\
    Conv1D \\
    \bottomrule
    \end{tabular}
    \label{tab:resnet_block1}
\end{minipage}
\hfill
\begin{minipage}{0.4\linewidth}
    \centering
    \caption{Basic Block Type 2}
    \begin{tabular}{@{}l@{}}
    \toprule
    \textbf{Layer} \\ \midrule
    BatchNorm \\
    ReLU \\
    Dropout \\
    Conv1D \\
    BatchNorm \\
    ReLU \\
    Dropout \\
    Conv1D \\
    Maxpool \\
    \bottomrule
    \end{tabular}
    \label{tab:resnet_block2}
\end{minipage}
\end{table}

In our downstream evaluation, logistic regression performed the best for depression detection using step count, whereas random forest is used for classifying flourishing using conversation duration. We obtained the best model using the following hyperparameters. Logistic regression: \{'penalty': ['l1', 'l2', 'elasticnet', 'none'], 'C': [0.1, 1, 10, 100], 'solver': ['saga'], 'l1\_ratio': [0.5, 0.7, 0.9]\} and Random Forest: \{'n\_estimators': [50, 100], 'max\_depth': [10, 20, 50], min\_samples\_split': [2, 10], 'min\_samples\_leaf': [1, 4], 'max\_features': ['sqrt', 'log2', None], 'bootstrap': [True, False]\}

\section{Data} \label{apd:data}

\subsection{MDD}

The MDD study recruited 300 participants for a 90-day study. The sample is predominantly middle-aged, with an average age of 40.13 years and an age range of 19 to 79 years, covering early adulthood to late older adulthood. The majority of participants are White (79\%), heterosexual (66\%), and women (84\%), with an average age of 40 years. The racial distribution of minority groups, including Black, Asian, and American Indian or Alaska Native participants, closely reflects the U.S. MDD population, while Hispanic and Latino individuals (12\%) match national demographic proportions. Most participants (93\%) have some college experience, with 68\% holding a college degree, and over half (61\%) are employed. Household incomes are distributed across all income brackets, largely aligning with the general U.S. distribution at the time of sampling. The study also maintained high engagement, with an average compliance rate of 82.07\%, based on daily Ecological Momentary Assessment (EMA) responses per participant over the 90-day period.

\begin{figure}[h]
    \centering
    \includegraphics[width=0.8\linewidth]{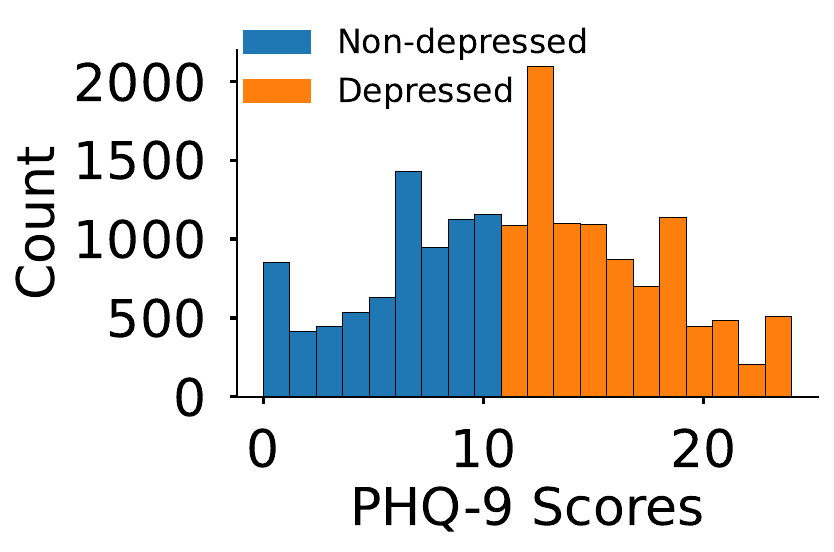}
    \caption{\textbf{Label statistics}. PHQ-9 Distribution in the MDD dataset. Scores $\ge$ 10 are considered depressed. 68\% of the days correspond to depression.}
    \label{fig:phq9}
\end{figure}

\section{Self-reported Surveys and Ecological Momentary Assessments}\label{apd:emas}

\subsection{Patient Health Questionnaire - 9 (PHQ-9) \citep{kroenke2001phq}}
In our MDD study, the user is asked over the past four hours if they have experienced the questions/statements in Table \ref{tab:phq9_scale}.

\begin{table*}[]
\centering
\caption{Patient Health Questionnaire - 9 (PHQ-9).}
\begin{tabular}{@{}ll@{}}
\toprule
\textbf{No.} &\textbf{Statement} \\
\midrule
1 &Little interest or pleasure in doing things \\
2 &Feeling down, depressed, or hopeless \\
3 &Trouble falling or staying asleep, or sleeping too much \\
4 &Feeling tired or having little energy \\
5 &Poor appetite or overeating \\
6 &Feeling bad about yourself or that you are a failure or
have let yourself or your family down \\
7 &Trouble concentrating on things, such as reading the
newspaper or watching television \\
8 &Moving or speaking so slowly that other people could
have noticed \\
9 &Thoughts that you would be better off dead, or of
hurting yourself \\
\bottomrule
\end{tabular} \label{tab:phq9_scale}
\end{table*}

\subsection{Flourishing Scale \citep{diener2010new}.} The user to indicate their agreement with each item in Table \ref{tab:flourishing_scale} by rating each statement from 1 to 7: (1) strongly disagree, (2) disagree, (3) slightly disagree, (4) mixed or neither agree nor disagree, (5) slightly agree, (6) agree, (7) strongly agree.

\begin{table*}[]
\centering
\caption{Flourishing scale statements.}
\begin{tabular}{@{}ll@{}}
\toprule
\textbf{No.} &\textbf{Statement} \\
\midrule
1 &I lead a purposeful and meaningful life. \\
2 &My social relationships are supportive and rewarding. \\
3 &I am engaged and interested in my daily activities \\
4 &I actively contribute to the happiness and well-being of others \\
5 &I am competent and capable in the activities that are important to me \\
6 &I am a good person and live a good life \\
7 &I am optimistic about my future \\
8 &People respect me \\
\bottomrule
\end{tabular} \label{tab:flourishing_scale}
\end{table*}

\section{Prompt Templates} \label{apd:prompt_templates}

\subsection{Depression Prompt}
The prompt template for depression is given below. In this template, we use temporal and health context. Furthermore, we convert the labels from 0 to 1 (as a probability). This approach produced better results than directly predicting 0 or 1. 

\begin{boxB}
    You are an AI assistant that evaluates if a user is depressed based on their step count in a day.\\

    Below is an instruction that describes a task, paired with an input that provides further context. 
    Write a response that appropriately completes the request in the format provided.\\

    Task: Some symptoms of depression are a lack of motivation and fatigue. It is linked to the number of steps taken in a day. Using the step count time series for a day, output a probability between 0 to 1 where 1 indicates high depression and 0 indicates low depression\\

    Input: 
        - Step Count: \{$<$"sensor data"$>$\}
\end{boxB}

\subsection{Depression Prompt with Knowledge}
The prompt template for depression is given below where add domain knowledge (highlighted).
\begin{boxB}
    You are an AI assistant that evaluates if a user is depressed based on their step count in a day.\\

    Below is an instruction that describes a task, paired with an input that provides further context. 
    Write a response that appropriately completes the request in the format provided.\\

    Task: Some symptoms of depression are a lack of motivation and fatigue. It is linked to the number of steps taken in a day \hl{such that higher step count is associated with lower depression symptoms.} Using the step count time series for a day, output a probability between 0 to 1 where 1 indicates high depression and 0 indicates low depression\\

    Input: 
        - Step Count: \{$<$"sensor data"$>$\}
\end{boxB}

\subsection{Flourishing Prompt} 
The prompt template for flourishing is given below. In this template, we use temporal and health context. 
\begin{boxB}
    You are an AI assistant that evaluates if a user is flourishing based on their conversation duration over time.\\

    Below is an instruction that describes a task, paired with an input that provides further context.
    Write a response that appropriately completes the request in the format provided.\\
    
    Task: We know that conversation duration is related to flourishing. Using the conversation duration (seconds) time series given below, output a probability between 0 to 1 where 1 indicates high flourishing and  0 indicates low flourishing. \\

    Input: 
        - Conversation duration: \{$<$"sensor data"$>$\}
\end{boxB}

\subsection{Flourishing Prompt with Domain Knowledge}
The prompt template for flourishing is given below. In this template, we use temporal and health context with domain knowledge.
\begin{boxB}
    You are an AI assistant that evaluates if a user is flourishing based on their conversation duration over time.\\

    Below is an instruction that describes a task, paired with an input that provides further context.
    Write a response that appropriately completes the request in the format provided.\\
    
    Task: We know that conversation duration is related to flourishing \hl{such that higher conversation duration is linked to higher flourishing.} Using the conversation duration (seconds) time series given below, output a probability between 0 to 1 where 1 indicates high flourishing and  0 indicates low flourishing. \\

    Input: 
        - Conversation duration: \{$<$"sensor data"$>$\}
\end{boxB}

\section{Efficiency Analysis} \label{apd:efficiency_analysis}
We compare the efficiency of Time2Lang and LlaMA prompting across two metrics: (1) \textbf{Latency}: the time taken for a single prediction and (2) \textbf{Throughput}: the number of model predictions per second (batch size = 16). From Table \ref{tab:efficiency}, we observe that Time2Lang demonstrates superior latency and throughput compared to LLaMA prompting, particularly as input sizes increase. Notably, while LLaMA prompting experiences a decline in efficiency with larger inputs, Time2Lang maintains near-constant latency and throughput across input sizes ranging from 512 to 4096, highlighting its scalability and efficiency.

\begin{table*}[h]
\caption{\textbf{Efficiency Analysis.} Latency and Throughput of Time2Lang and LLaMA prompting at different input sizes.}
\centering
\begin{tabular}{@{}lllll@{}}
\toprule
\multirow{2}{*}{\textbf{Input Size}} & \multicolumn{2}{c}{\textbf{Latency} (ms) $\downarrow$} & \multicolumn{2}{c}{\textbf{Throughput} (predictions per second) $\uparrow$} \\ \cmidrule(l){2-5} 
                    & Time2Lang           & LLaMA          & Time2Lang            & LLaMA            \\ \midrule 
256 &30.5 (0.1) &314.2 (2.1) &70.07 (0.25) &15.19 (0.27) \\
512 &50.4 (0.2) &341.8 (1.0) & 33.69 (0.25) &8.70 (0.17) \\
1024 &50.6 (0.2) &414.1 (1.0) &33.43 (0.30) &4.59 (0.08) \\
2048 &50.7 (0.2) &572.5 (2.2) &32.85 (0.28) &2.28 (0.04) \\
4096 &50.9 (0.1) &1016.5 (2.5) &32.69 (0.21) &1.05 (0.08)
                 \\ \bottomrule
\end{tabular} \label{tab:efficiency}
\end{table*}

\section{Comparison to Supervised Task-Specific Models}
Here, we train two different ResNets specifically to predict depression and flourishing from step count and conversation duration, respectively. However, it is important to note that \textbf{Time2Lang ($f$ and $g$) is not pre-trained on downstream datasets.} Instead, it relies on feature extraction followed by a random forest classifier. This fundamental difference in training methodology makes direct comparison unfair, as task-specific models benefit from direct supervision on the downstream data while Time2Lang does not. Nevertheless, this experiment offers valuable insights into the trade-offs between task-specific deep learning and general-purpose reprogramming.

From Table \ref{tab:fully_supervised}, we observe that the task-specific models achieve better performance in three out of four cases. Specifically, for depression prediction, the task-specific models attains an AUROC of 0.60 and an AUPRC of 0.72, whereas Time2Lang achieves an AUROC of 0.57 and a slightly higher AUPRC of 0.73. In the case of flourishing prediction, the task-specific model outperforms Time2Lang in both metrics, with an AUROC of 0.75 compared to 0.71 and an AUPRC of 0.82 compared to 0.74.

\begin{table*}[] 
\centering
\caption{\textbf{Task-specific deep learning baseline}. Higher values indicate better performance.}
\begin{tabular}{lllll}
\toprule
\multirow{2}{*}{\textbf{Model}} & \multicolumn{2}{c}{\textbf{Depression (step count)}} & \multicolumn{2}{c}{\textbf{Flourishing (conversation)}} \\ \cmidrule{2-5}
    & AUROC                 & AUPRC                & AUROC                  & AUPRC                 \\ \midrule
Task-specific ResNet &\textbf{0.60} (0.02) & 0.72 (0.01) & \textbf{0.75} (0.03) &\textbf{0.82} (0.03)  \\
 Time2Lang &0.57 (0.01) &\textbf{0.73} (0.00) & 0.71 (0.10) &0.74 (0.11) \\
 \bottomrule
\end{tabular} \label{tab:fully_supervised}
\end{table*}

\section{Additional Ablation Studies} \label{apd:ablation}

\subsection{Effect of residual connection}\label{apd:residual_ablation}
We assess the effect of incorporating residual connections during training. As shown in Table \ref{tab:residual}, the model with the residual connection performed better, showing improvements in three out of four cases. For depression detection using step count, AUROC increases from 0.54 to 0.57, and AUPRC improves from 0.70 to 0.73. Similarly, in the flourishing prediction task based on conversation data, AUROC increases slightly from 0.70 to 0.71, while AUPRC decreases from 0.77 to 0.74. These results indicate that residual connections improve performance, particularly in depression detection, by preserving and effectively propagating temporal patterns. Additionally, the model with residual connections consistently achieves higher AUROC scores for both depression detection (0.57 vs. 0.54) and flourishing prediction (0.71 vs. 0.70), suggesting improved discriminative power in distinguishing between positive and negative cases.

\begin{table*}[] 
\centering
\caption{\textbf{Time2Lang residual ablation}. Evaluating the impact of the residual connection. Higher values indicate better performance.}
\begin{tabular}{lllll}
\toprule
\multirow{2}{*}{\textbf{Model}} & \multicolumn{2}{c}{\textbf{Depression (step count)}} & \multicolumn{2}{c}{\textbf{Flourishing (conversation)}} \\ \cmidrule{2-5}
    & AUROC                 & AUPRC                & AUROC                  & AUPRC                 \\ \midrule
 Time2Lang (w/o residual) &0.54 (0.01) &0.70 (0.01) &0.70 (0.15) &\textbf{0.77} (0.13) \\
 Time2Lang (w/ residual) &\textbf{0.57} (0.01) &\textbf{0.73} (0.00) & \textbf{0.71} (0.10) &0.74 (0.11) \\
 \bottomrule
\end{tabular} \label{tab:residual}
\end{table*}

\subsection{Architecture ablations}\label{apd:architecture_ablation}
To analyze how different choices in input encoder and projection architecture impact Time2Lang, we conduct an architecture study as follows. The default setup uses a ResNet as the base encoder and a projection module composed of two fully connected layers, as detailed in Appendix~\S \ref{apd:architecture}. We examine three architectural variations: increasing the base filter size to 64 for a wider encoder, adding more residual blocks for a deeper encoder, and expanding the projection embedding size from 256 to 512. The resulting changes in model size and downstream performance are presented in Table \ref{tab:architecture_ablations}.

From Table \ref{tab:architecture_ablations}, we observe that the best overall performance is achieved by wider models with 64 base filters and a larger embedding size of 512. This configuration results in the highest AUPRC scores for both depression (0.73) and flourishing (0.85), and the highest AUROC for flourishing (0.80). In contrast, increasing the model depth by using 10 residual blocks does not lead to consistent performance improvements. In fact, deeper models tend to perform slightly worse or similarly across both embedding sizes, suggesting that depth does not significantly benefit Time2Lang's downstream performance in this setup.

\begin{table*}[]
\centering
\caption{\textbf{Time2Lang architecture ablation.} Evaluating the impact of input encoders ($f$) and projection ($g$) architectures on downstream performance.}
\scalebox{0.80}{
\begin{tabular}{@{}lllllllll@{}}
\toprule
\multirow{2}{*}{\textbf{Base filters}} & \multirow{2}{*}{\textbf{Blocks}} & \multirow{2}{*}{\textbf{Embedding Size}} & \multirow{2}{*}{\textbf{$f$ parameters}} & \multirow{2}{*}{\textbf{$g$ parameters}} & \multicolumn{2}{c}{\textbf{Depression}} & \multicolumn{2}{c}{\textbf{Flourishing}} \\
\cmidrule{6-9}
& & & & &AUROC & AUPRC & AUROC & AUPRC \\
\midrule
32 &6 &256 &300K &1.8M &0.57 (0.01) & 0.73 (0.00) &0.71 (0.10) &0.74 (0.11) \\
64 &6 &256 &600K &1.8M &0.55 (0.00) &0.72 (0.00) &0.31 (0.11) & 0.47 (0.08) \\
64 &10 &256 &1.5M &1.8M &0.55 (0.00) &0.71 (0.00) &0.67 (0.10) &0.72 (0.08) \\
32 &6 &512 &300K &2M &0.57 (0.00) &0.73 (0.00) &0.37 (0.13) &0.53 (0.13) \\
64 &6 &512 &600K &2M &0.57 (0.00) &0.73 (0.00) &0.80 (0.06) &0.85 (0.05) \\
64 &10 &512 &1.5M &2M &0.54 (0.00) &0.71 (0.00) &0.42 (0.17) &0.59 (0.11) \\
                                       \bottomrule
\end{tabular}} \label{tab:architecture_ablations}
\end{table*}

\subsection{LLM ablation}\label{apd:llm_ablation}
We evaluate different LLM backbones in the Time2Lang framework (Table \ref{tab:llm_backbones}). In particular, we considered LLaMA 3.2 \citep{dubey2024llama}, Gemma 3 \citep{team2025gemma}, and OLMo \citep{groeneveld2024olmo} with approximately 1B parameters. For depression prediction, all models achieved comparable performance, with AUROC values ranging from 0.57 to 0.58 and AUPRC consistently at 0.73 across all backbones. Gemma 3 slightly outperformed the others in AUROC. For flourishing prediction, LLaMA 3.2 substantially outperformed both Gemma 3 and OLMo, achieving an AUROC of 0.71 and AUPRC of 0.74, suggesting stronger capability in extracting meaningful representations from conversational data. In contrast, Gemma 3 performed the worst in this task, with an AUROC of 0.41 and an AUPRC of 0.48. Perhaps, Gemma's decoder embedding size of 1152 is not compatible with the projection $g$ or the TFM. Future work should include a broader comparison across architectures and hyperparameter settings to support more definitive conclusions. OLMo offered modest improvements over Gemma 3 but was still considerably behind LLaMA 3.2. Overall, while all backbones performed similarly on the step count-based depression task, LLaMA 3.2 demonstrated a significant advantage in the more flourishing prediction task. 

\begin{table*}[] 
\centering
\caption{\textbf{Time2Lang with different LLM backbones}. All LLMs have $\sim$1B parameters. Higher values indicate better performance.}
\begin{tabular}{lllll}
\toprule
\multirow{2}{*}{\textbf{LLM}} & \multicolumn{2}{c}{\textbf{Depression (step count)}} & \multicolumn{2}{c}{\textbf{Flourishing (conversation)}} \\ \cmidrule{2-5}
    & AUROC                 & AUPRC                & AUROC                  & AUPRC                 \\ \midrule
 Gemma 3 &0.58 (0.06) &0.73 (0.00) & 0.41 (0.06) &0.48 (0.03) \\
 OLMo &0.57 (0.00) &0.73 (0.00) &0.49 (0.06) &0.57 (0.08) \\
 LLaMA 3.2 &0.57 (0.01) &0.73 (0.00) & 0.71 (0.10) &0.74 (0.11) \\
 \bottomrule
\end{tabular} \label{tab:llm_backbones}
\end{table*}

\section{Extended Discussion}

In this paper, we assess the capabilities of Time2Lang for mental health sensing. However, the broader concepts introduced can be applied to various sensing fields, multi-modal learning, and explainable AI. Specifically, in healthcare, analyzing physiological signals such as PPG and ECG could yield valuable insights into cardiovascular irregularities. For instance, \citet{cai2024jolt} integrate ECG and text data into a shared representation to summarize ECG readings. Similarly, for PPG analysis, the Time2Lang TFM encoder could potentially be replaced with a modality-specific encoder like PaPaGei \citep{pillai2024papagei}. Future studies could explore additional health-related time series data, including heart rate, respiratory patterns, and ballistocardiography.

Time2Lang employs Chronos as its TFM and LlaMA as its LLM. Advancing research in this area will benefit from systematic analysis of the choice of TFM and LLM in reprogramming frameworks. Furthermore, healthcare data is multi-modal consisting of images, audio, text, and sensors. Unifying these modalities through an LLM remains challenging. We envision extending Time2Lang to investigate methods to integrate these diverse data types using reprogramming techniques. A potential approach involves using specialized encoders for each modality and merging them within the reprogramming framework.

Missing data is a common challenge in real-world sensor-based datasets. Since Time2Lang is trained on synthetic data, missing values in downstream datasets are addressed through a two-step process: first, by computing the mean via downsampling within a defined window, and second, by zero-filling the remaining gaps. Future research can explore many different imputation strategies \citep{bhattacharya2024imputation}. More importantly, we envision improving Time2Lang’s resilience to missing data by adopting a masked autoencoder-style architecture, enabling the model to learn how to reconstruct missing segments during training. Additionally, real-world healthcare data often contains missing modalities that may not be perfectly synchronized. As a result, model architecture and training methodologies must be designed to effectively manage both absent and misaligned data sources. For instance, CroSSL \citep{deldari2024crossl} addresses these challenges by implementing strategies to enhance robustness in such scenarios.

Understanding how TFMs and LLMs interact is essential for interpreting how large models learn temporal patterns and generate outputs. Mechanistic interpretability techniques such as probing activation layers and conducting layer-wise attribution analyses could potentially offer insights for combining different FMs \citep{conmy2023towards}. Furthermore, counterfactual explanations and concept-based interpretability approaches could enhance transparency in learned embeddings, a crucial aspect in healthcare applications where explainability is paramount.

\end{document}